\documentclass[letterpaper]{article} 
\usepackage{aaai2026}  
\usepackage{times}  
\usepackage{helvet}  
\usepackage{courier}  
\usepackage[hyphens]{url}  
\usepackage{graphicx} 
\usepackage{natbib}  
\usepackage{caption} 
\usepackage{algorithm}
\usepackage{amsmath}  
\usepackage{amssymb}  
\usepackage{amsfonts} 
\usepackage{bm}       
\usepackage{dsfont}
\usepackage{booktabs} 
\usepackage{multirow}   
\usepackage{array}      
\usepackage{graphicx}   
\usepackage{caption}    
\usepackage{siunitx}    
\usepackage{xcolor}     
\usepackage[table]{xcolor}
\usepackage{etoolbox}   
\usepackage{cleveref}
\usepackage{subcaption}
\usepackage{rotating}
\usepackage[utf8]{inputenc}
\usepackage{algpseudocode} 
\usepackage{threeparttable}
\usepackage{amsthm}
\usepackage{enumitem}

\usepackage{newfloat}
\usepackage{listings}
\DeclareCaptionStyle{ruled}{labelfont=normalfont,labelsep=colon,strut=off} 
\floatstyle{ruled}
\newfloat{listing}{tb}{lst}{}
\floatname{listing}{Listing}
\title{UCPO: A Universal Constrained Combinatorial Optimization Method via Preference Optimization}
\author{
    Zhanhong Fang\textsuperscript{\rm 1}, Debing Wang\textsuperscript{\rm 1}, Jinbiao Chen\textsuperscript{\rm 1}, Jiahai Wang\textsuperscript{\rm 1}, Zizhen Zhang\textsuperscript{\rm 1}\thanks{Corresponding author.}
}
\affiliations{
    \textsuperscript{\rm 1}Sun Yat-Sen University\\

}

\usepackage{bibentry}

\begin{document}

\maketitle

\begin{abstract}
Neural solvers have demonstrated remarkable success in combinatorial optimization, often surpassing traditional heuristics in speed, solution quality, and generalization. However, their efficacy deteriorates significantly when confronted with complex constraints that cannot be effectively managed through simple masking mechanisms. To address this limitation, we introduce Universal Constrained Preference Optimization (UCPO), a novel plug-and-play framework that seamlessly integrates preference learning into existing neural solvers via a specially designed loss function, without requiring architectural modifications. UCPO embeds constraint satisfaction directly into a preference-based objective, eliminating the need for meticulous hyperparameter tuning. Leveraging a lightweight warm-start fine-tuning protocol, UCPO enables pre-trained models to consistently produce near-optimal, feasible solutions on challenging constraint-laden tasks, achieving exceptional performance with as little as 1\% of the original training budget.
\end{abstract}


\section{Introduction}

Combinatorial optimization forms the backbone of numerous mission-critical applications across modern industry and scientific computing. Despite the exponential complexity of their search spaces, these tasks demand near-optimal solutions within practical timeframes. Recent advancements in neural solvers, particularly those grounded in deep reinforcement learning (DRL), have demonstrated significant superiority over classical heuristics in terms of solution quality, computational efficiency, and cross-domain generalization. These solvers have achieved remarkable performance on canonical benchmarks such as the Traveling Salesman Problem (TSP), Capacitated Vehicle Routing Problem (CVRP), and Job-Shop Scheduling Problem (JSP). However, a critical challenge arises when deploying these methods on real-world instances with complex constraints that cannot be effectively managed through simple masking mechanisms.

In practical settings, hard constraints, e.g., non-negotiable customer time windows in vehicle routing \cite{su151512004} or draft-depth limitations in port logistics \cite{FADDA2023106024}, may pose significant difficulties. Traditional neural solvers often struggle to balance feasibility and optimality under such constraints. Existing approaches either rely on intricate masking strategies \cite{chen2024looking}, which become computationally infeasible as the decision horizon grows, or employ Lagrangian penalty functions \cite{tang2022learning, bi2024learning}, which transform the original single-objective problem into a bi-objective trade-off. The resulting Lagrange multipliers are notoriously sensitive to calibration, often leading to solutions that are either infeasible or suboptimal. This sensitivity necessitates careful, problem-specific tuning, which is both time-consuming and limits the universal applicability of these methods.

To address these limitations, we introduce \textbf{Universal Constrained Preference Optimization (UCPO)}, a novel framework that leverages preference optimization \cite{meng2024simpo,pan2025preference,liao2025bopo} to handle hard constraints in combinatorial optimization tasks.

Our framework introduces the following key innovations:
\begin{enumerate}
    \item \textbf{Seamless Architectural Integration}: UCPO can be effortlessly attached to existing neural combinatorial optimization (NCO) models without altering the underlying architecture. We design a novel universal constrained preference loss function, which includes feasibility margin loss, primal refinement loss, and dual exploration loss. This approach preserves the original model’s learned representations while significantly enhancing its ability to handle complex constraints.
    \item \textbf{Mask-Agnostic Constraint Enforcement}: Unlike traditional methods that rely on progressively refined or multi-step masks to ensure feasibility, UCPO employs a unified partial-order criterion. This criterion prioritizes feasible solutions over infeasible ones and favors infeasible solutions with lower aggregate constraint violations. It also eliminates the need for hand-tuned or adaptively updated Lagrange multipliers of the relaxation problem.
    \item \textbf{Efficient Warm-Start Adaptation}: Utilizing a pre-trained checkpoint as a high-quality initializer, UCPO performs lightweight fine-tuning through pairwise preference sampling and gradient accumulation. This protocol significantly reduces computational resource requirements and enables rapid adaptation to new constraint sets without retraining from scratch.
\end{enumerate}

We conducted extensive experiments across four representative constrained benchmarks: Traveling Salesman Problem with Time Windows (TSPTW), Capacitated Vehicle Routing Problem with Time Windows (CVRPTW), Traveling Salesman Problem with Draft Limit (TSPDL), and Capacitated Vehicle Routing Problem with Time Windows and Limited Vehicles (CVRPTWLV). Each benchmark involves intricate constraints that are notoriously difficult to handle, such as time-windows and resource limitations. UCPO demonstrates exceptional performance compared to state-of-the-art neural methods, establishing it as a general, reusable, and highly efficient paradigm for handling hard constraints in neural combinatorial optimization. 

\section{Related Works}

Neural solvers for combinatorial optimization can be broadly categorized into three principal classes: end-to-end constructive methods, iterative improvement methods, and non-autoregressive methods. This work primarily focuses on end-to-end constructive methods.

\begin{itemize}[leftmargin=0pt]
    \item[] \textbf{End-to-end constructive methods.} Initiated by Pointer Networks \cite{vinyals2015pointer}, this category generates solutions step-by-step using autoregressive decoding. Subsequent works such as the Attention Model (AM) \cite{kool2018attention} and H-TSP \cite{pan2023h} enhance policy expressiveness through reinforcement learning and hierarchical decoding. POMO \cite{kwon2020pomo} achieves a significant performance leap in small-scale TSP by leveraging solution space symmetries to improve sample efficiency. SymNCO \cite{kim2022sym} adopts a similar strategy. MatNet \cite{kwon2021matrix}, ELG \cite{gao2023towards}, and PolyNet \cite{hottung2024polynet} further improve accuracy for small instances through matrix distance calculations, local policy optimization, and solution diversity mechanisms. To handle large-scale problems, LEHD \cite{luo2023neural} and InViT \cite{fang2024invit} enhance representation capacity and decoding efficiency. GOAL \cite{drakulic2025goal}, MTPOMO \cite{liu2024multi}, and MVMoE \cite{zhou2024mvmoe} pursue multi-task generalization. Notably, BOPO \cite{liao2025bopo} and POCO \cite{pan2025preference} introduce preference-optimization concepts, offering novel policy alignment strategies. However, these methods primarily rely on lightweight masking mechanisms, necessitating ad-hoc penalty functions or masking logics when handling additional constraints, thereby limiting their universality.

    \item[] \textbf{Other paradigms.} Iterative improvement methods (e.g., DACT \cite{ma2021learning}, NeuOpt \cite{ma2023learning}) improve solutions via refinement loops, while non-autoregressive methods (e.g., DeepACO \cite{ye2023deepaco}, DIFUSCO \cite{sun2023difusco}) aim to enhance inference efficiency through ant colony or diffusion-based generation. Notably, these paradigms also struggle with global constraints under the same masking limitations.
\end{itemize}

When constraints are global, non-linear, or both, lightweight masking is often insufficient. Existing work tackles this via two main strategies.
\begin{itemize}[leftmargin=0pt]
    \item[] \textbf{Lagrangian relaxation and penalty functions.} A common strategy involves augmenting the objective function with soft constraint penalties or employing two-stage training to mitigate constraint violations \cite{tang2022learning, chen2022deep, bi2024learning}. For instance, MUSLA \cite{chen2024looking} incorporates multi-step look-ahead features but is coupled to problem-specific characteristics, limiting its flexibility. PIP \cite{bi2024learning} combines Lagrange multipliers with one-step look-ahead masks to reduce infeasibility rates but struggles to balance feasibility and solution quality. These methods require careful tuning of Lagrange multipliers, which is sensitive to problem-specific parameters and may lead to suboptimal solutions.
    \item[] \textbf{Exploration of feasible and infeasible regions.} NeuOpt \cite{ma2023learning} guides exploration into infeasible regions followed by $k$-opt iterative repair. NCG \cite{Xia2024ANC} adopts a column-generation paradigm to predict the probability of constraint satisfaction for each column. Both approaches require increased exploration budgets or the training of auxiliary probabilistic models, adding to the computational and implementation complexities.
\end{itemize}

In summary, existing methods for combinatorial optimization face challenges in handling complex constraints, often requiring problem-specific tuning and complex masking strategies. This motivates the need for a versatile, mask-free, and low-hyperparameter solution compatible with general-purpose neural solvers. Our UCPO framework addresses these limitations by offering an efficient and universal approach to constrained combinatorial optimization.

\begin{figure*}[t]
\centering
\includegraphics[scale=0.45]{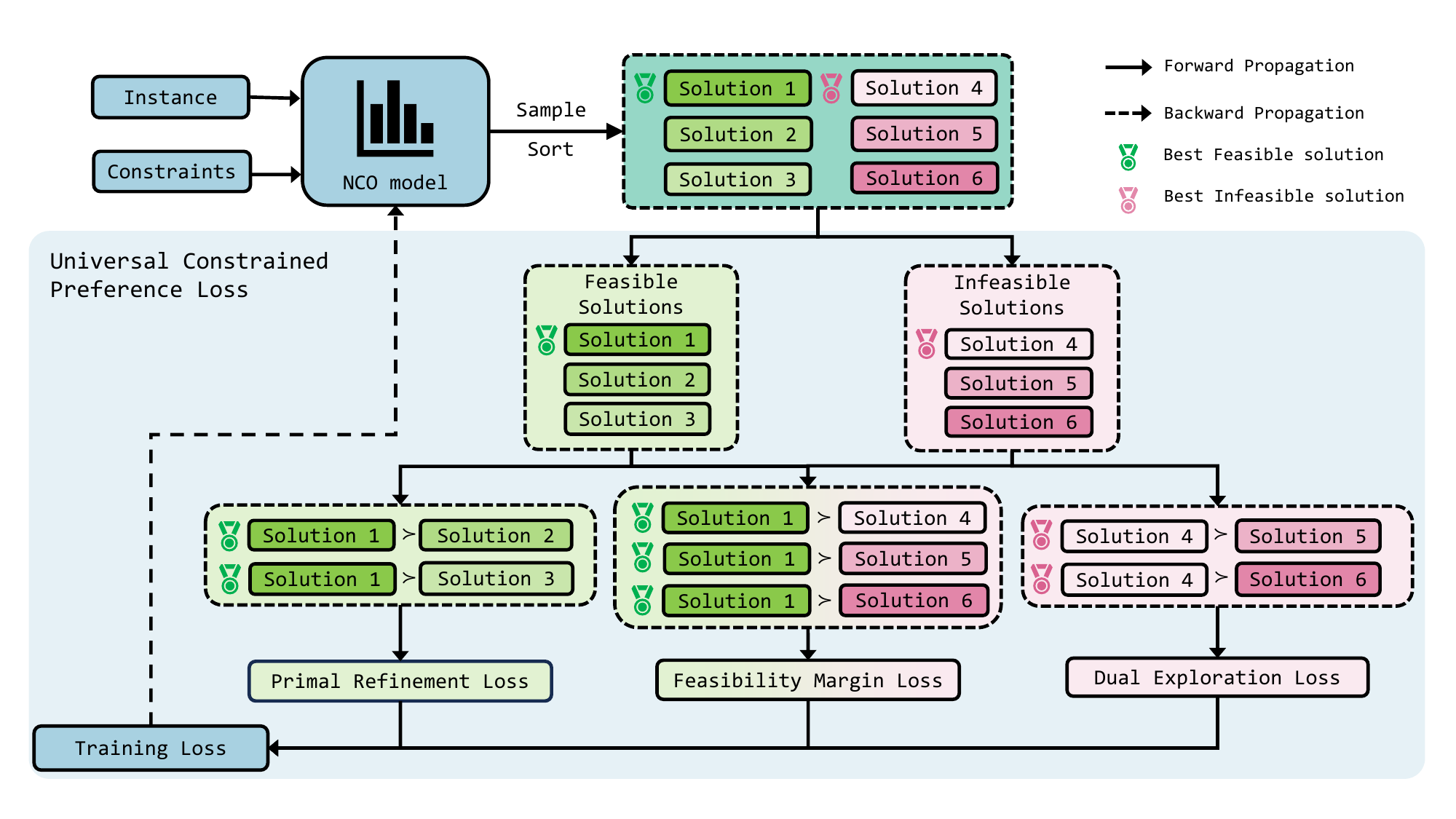}
\caption{Overview of the Universal Constrained Preference Optimization (UCPO) framework. The process begins by sampling candidate solutions from the base NCO model, followed by fine-tuning the model using Universal Constrained Preference Loss.}
\label{fig1}
\end{figure*}

\section{Preliminaries}

\subsection{Lagrangian Methods}

Consider a generic combinatorial optimization problem of the form:
\[
\begin{aligned}
\min_{x\in\mathcal{X}} & \quad f(x) \\
\text{subject to} & \quad g_j(x) \leq 0, \quad j \in \mathcal{J}, \\
& \quad h_k(x) = 0, \quad k \in \mathcal{K},
\end{aligned}
\]
where \(x = (x_1, \dots, x_n)^\top\) represents the decision variables, \(\mathcal{X} \subseteq \mathbb{R}^n\) defines the feasible domain, \(f(x)\) is the objective function to be minimized, and \(g_j(x)\) and \(h_k(x)\) represent inequality and equality constraints, respectively. Detailed mathematical formulations for the specific problems addressed in this paper are provided in the Appendix A.

To handle constraints, we introduce non-negative Lagrange multipliers \(\lambda = (\lambda_j)_{j \in \mathcal{J}} \geq 0\) and unconstrained multipliers \(\mu = (\mu_k)_{k \in \mathcal{K}} \in \mathbb{R}\), and construct the Lagrangian:
\[
\mathcal{L}(x, \lambda, \mu) = f(x) + \sum_{j \in \mathcal{J}} \lambda_j g_j(x) + \sum_{k \in \mathcal{K}} \mu_k h_k(x).
\]

The dual function is defined as:
\[
\theta(\lambda, \mu) = \inf_{x \in \mathcal{X}} \mathcal{L}(x, \lambda, \mu).
\]

To reduce computational complexity, the multipliers are often fixed at certain values \((\bar{\lambda}, \bar{\mu})\), transforming the dual problem into a single-level unconstrained optimization problem:
\[
\min_{x \in \mathcal{X}} \left\{ f(x) + \bar{\lambda}^\top g(x) + \bar{\mu}^\top h(x) \right\}.
\]

This approach converts the original saddle-point problem over \((x, \lambda, \mu)\) into a single-variable minimization problem over \(x \in \mathcal{X}\). As a result, any off-the-shelf unconstrained or model-free algorithm (e.g., REINFORCE \cite{williams1992simple}) can be directly applied without the need for an outer loop to update the multipliers. However, this simplicity comes at the cost of losing strict feasibility guarantees unless \((\bar{\lambda}, \bar{\mu})\) corresponds to an optimal dual solution. In practice, the multipliers are treated as tunable ``penalty coefficients", enabling the generation of approximately feasible solutions through manual or heuristic adjustments.

\subsection{Preference Optimization Methods}

Given an instance \(x \in \mathcal{X}\) and a policy network with parameters \(\theta\), the policy \(\pi_\theta(\cdot \mid x)\) generates \(N\) candidate solutions:
\[
\mathcal{T}_x = \{\tau_i\}_{i=1}^N \sim \pi_\theta(\cdot \mid x).
\]

For every ordered pair \((\tau_i, \tau_j)\), we define the preference label as:
\[
y_{ij} = \mathds{1}\!\left\{f(x,\tau_i) < f(x,\tau_j)\right\},
\]
where \(f(x,\tau)\) denotes the objective value of solution \(\tau\) on instance \(x\). The symbol \(\mathds{1}\) indicates that \(\tau_i\) is strictly better than \(\tau_j\).

The Bradley–Terry model \cite{bradley1952rank} translates the objective gap into a preference probability:
\[
p_\theta(\tau_i \succ \tau_j \mid x) = \sigma\!\bigl(\beta \cdot d_\theta(x,\tau_i,\tau_j)\bigr),
\]
where \(\sigma(\cdot)\) is the logistic (sigmoid) function, \(d_\theta(x,\tau_i,\tau_j) = \alpha\bigl[\log \pi_\theta(\tau_i \mid x) - \log \pi_\theta(\tau_j \mid x)\bigr]\) represents the log-likelihood difference, \(\alpha > 0\) is a temperature parameter, and \(\beta\) is an adaptive scaling factor (BOPO \cite{liao2025bopo} sets \(\beta = f(x,\tau_j)/f(x,\tau_i)\), while POCO \cite{pan2025preference} simply fixes \(\beta = 1\)).

Maximizing the likelihood over all preference pairs is equivalent to minimizing the loss:
\[
\mathcal{L}(\theta) = -\mathds{E}_{x\sim\mathcal{X}} \frac{1}{|\mathcal{T}_x|^2} \sum_{i \neq j} y_{ij} \log p_\theta(\tau_i \succ \tau_j \mid x).
\]

Substituting the Bradley–Terry model yields the closed-form objective:
\[
\begin{aligned}
\mathcal{L}(\theta) = -\mathds{E}_{x\sim\mathcal{X}} \frac{1}{|\mathcal{T}_x|^2} \sum_{i \neq j} y_{ij} \, \log \sigma\!\Bigl( & \alpha \beta \bigl[ \log \pi_\theta(\tau_i \mid x) \\
& - \log \pi_\theta(\tau_j \mid x) \bigr] \Bigr).
\end{aligned}
\]

\section{Methodology}

To address complex constrained combinatorial optimization problems, we introduce Universal Constrained Preference Optimization (UCPO), a novel framework that seamlessly integrates preference learning into the combinatorial optimization workflow in a plug-and-play manner. Designed as a lightweight loss function, UCPO can be easily incorporated into any existing RL-based solver without the need for extensive architectural modifications or complete retraining. This is achieved by leveraging a warm-start fine-tuning procedure that utilizes existing pre-trained checkpoints, thereby significantly enhancing performance in constrained settings with minimal additional computational overhead.

The overall architecture of UCPO is illustrated in Figure \ref{fig1}. The detailed components are presented as follows.

\subsection{Constrained Preference Optimization Formulation}

\subsubsection{Partial-Order Relation.}
Consider an instance \(x \in \mathcal{X}\) with constraint sets \(\mathcal{G} = \{g_j(x) \leq 0, j \in \mathcal{J}\}\) and \(\mathcal{H} = \{h_k(x) = 0, k \in \mathcal{K}\}\). The policy \(\pi_\theta(\cdot \mid x)\) generates \(N\) candidate solutions \(\mathcal{T}_x = \{\tau_i\}_{i=1}^N\), some of which may violate the constraints. To address infeasible solutions, we introduce the indicator function:
\[
I(x,\tau)=\begin{cases}
1,& \exists\, j\in\mathcal{J}: g_j(x,\tau)>0 \text{ or }\\
  &\exists\, k\in\mathcal{K}: h_k(x,\tau)\neq 0,\\
0,& \forall\, j\in\mathcal{J}: g_j(x,\tau)\le 0 \text{ and }\\
  & \forall\, k\in\mathcal{K}: h_k(x,\tau)=0.
\end{cases}
\]

Here, \(I(x, \tau) = 0\) if and only if \(\tau\) is strictly feasible. Using the quadruple \((\tau_i, \tau_j, I(x, \tau_i), I(x, \tau_j))\), we define the partial order \(\tau_i \succ \tau_j\) if exactly one of the following conditions holds:
\[
\begin{cases}
I(x, \tau_i) = I(x, \tau_j) = 0 \text{ and } f(x, \tau_i) < f(x, \tau_j), \\
I(x, \tau_i) = 0 \text{ and } I(x, \tau_j) = 1, \\
I(x, \tau_i) = I(x, \tau_j) = 1 \text{ and } L(x, \tau_i, \lambda, \mu) < L(x, \tau_j, \lambda, \mu),
\end{cases}
\]
where 
\[
L(x, \tau, \lambda, \mu) = f(x, \tau) + \sum_{j \in \mathcal{J}} \lambda_j g_j(x, \tau) + \sum_{k \in \mathcal{K}} \mu_k h_k(x, \tau)
\]
is the Lagrangian with non-negative multipliers \(\lambda = (\lambda_j)_{j \in \mathcal{J}} \geq 0\) and unconstrained multipliers \(\mu = (\mu_k)_{k \in \mathcal{K}} \in \mathbb{R}\). This hierarchy prioritizes: 1) lower objective values among feasible solutions, 2) feasible solutions over infeasible ones, and 3) lower Lagrangian values among infeasible solutions. Empirical validation in the Appendix confirms this strategy's superiority.

\subsubsection{Bradley–Terry Model.}
Following the typical formalization of preference optimization, we convert the partial order into a preference probability. For a solution pair \((\tau_i, \tau_j)\) and instance \(x\), the preference probability is defined as:
\[
p_\theta(\tau_i \succ \tau_j \mid x) = \sigma\!\left(\beta \left[\log \pi_\theta(\tau_i \mid x) - \log \pi_\theta(\tau_j \mid x)\right]\right),
\]
where \(\log \pi_\theta(\tau \mid x)\) is the log-likelihood of \(\tau\) under the policy \(\pi_\theta\), \(\sigma(\cdot)\) is the logistic sigmoid function, and \(\beta\) is an adaptive scaling factor. Following the parameterization adopted in BOPO \cite{liao2025bopo}, we dispense with the temperature coefficient \(\alpha\) and retain only the scalar \(\beta\). This formulation efficiently captures the relative preference between solutions while reducing computational complexity.

\subsection{Design of Universal Constrained Preference Loss}

To address environments with complex constraints and sparse feasible solutions, we design three loss components that progressively guide the policy from infeasible exploration to high-quality feasible solution learning. Although not strictly mutually exclusive, these losses follow a natural progression: \textit{Dual Exploration}~$\rightarrow$~\textit{Feasibility Margin}~$\rightarrow$~\textit{Primal Refinement}, which enables the model to balance constraint satisfaction with objective optimality in a principled manner.

Given a problem instance and its constraints, the policy network concurrently samples \(N\) complete trajectories. While this yields \(\binom{N}{2}\) pairwise preferences in theory, we instead enforce the previously established partial order to construct a \emph{single ranked list}:
\[
\tau_1\succ\tau_2\succ\dots\succ\tau_N,
\]  
which immediately partitions the candidates into  
\[
\begin{aligned}
\mathcal T^{\!T}=\{\tau\in\mathcal T_x\mid I(x,\tau)=0\}\quad &\text{(feasible)},\\
\mathcal T^{\!F}=\{\tau\in\mathcal T_x\mid I(x,\tau)=1\}\quad &\text{(infeasible)}.
\end{aligned}
\]  

Either subset may be empty, especially during early training stages. Our loss design guides the policy toward strict feasibility while refining objective quality by activating three terms depending on the realization of \(\mathcal T^{\!T}\) and \(\mathcal T^{\!F}\).

\paragraph{Dual Exploration Loss \(\mathcal L^{\text{Dual}}\).}  
Activated only when \(\mathcal T^{\!T}=\emptyset\), i.e., all sampled trajectories are infeasible. Let \(\tau_\circ=\arg\min_{\tau\in\mathcal T^{\!F}}L(x,\tau,\lambda,\mu)\) be the ``least-infeasible" solution under particular Lagrange multiplier. We contrast \(\tau_\circ\) against the remaining infeasible trajectories to guide exploration toward lower constraint violation:
\begin{equation}
\begin{aligned}
\mathcal{L}^{\text{Dual}}(\theta)=
  &-\mathbb{E}_{x\sim\mathcal{X}}
\frac{1}{|\mathcal{T}^{\!F}|-1}\sum_{\tau\in\mathcal{T}^{\!F}\setminus\{\tau_\circ\}}
\log\sigma\!\Bigl(\\
  &\beta^{\text{Dual}}\bigl[\log\pi_\theta(\tau_\circ\!\mid\!x)-\log\pi_\theta(\tau\!\mid\!x)\bigr]\Bigr),
\end{aligned}
\end{equation}
with \(\beta^{\text{Dual}}=L(x,\tau,\lambda,\mu)/L(x,\tau_\circ,\lambda,\mu)\ge 1\). Alternative pairing strategies are analyzed in Appendix D.

It is worth noting that \(\mathcal{L}^{\text{Dual}}\) is activated only during the very beginning of training, while no feasible solution has yet been found. Once feasible solutions are sampled (via exploration guided by the Lagrangian-based ranking), the loss is deactivated in favor of \(\mathcal{L}^{\text{Margin}}\) and \(\mathcal{L}^{\text{Primal}}\), which operate solely within the feasible region and are multiplier-agnostic. Consequently, the long-term policy behavior is dominated by constraint-compliant optimization, decoupling final performance from the initial \((\lambda, \mu)\).  

\paragraph{Feasibility Margin Loss \(\mathcal L^{\text{Margin}}\).}  
Activated when both \(\mathcal T^{\!T}\) and \(\mathcal T^{\!F}\) are non-empty. Let \(\tau_\star=\arg\min_{\tau\in\mathcal T^{\!T}}f(x,\tau)\) be the best feasible solution. We contrast \(\tau_\star\) against every infeasible trajectory \(\tau\in\mathcal T^{\!F}\):
\begin{equation}
\begin{aligned}
\mathcal{L}^{\text{Margin}}&(\theta)=
  -\mathbb{E}_{x\sim\mathcal{X}}
\frac{1}{|\mathcal{T}^{\!F}|}\sum_{\tau\in\mathcal{T}^{\!F}}
\log\sigma\!\Bigl( \\
  &\beta^{\text{Margin}}\bigl[\log\pi_\theta(\tau_{r_\star}\!\mid\!x)-\log\pi_\theta(\tau\!\mid\!x)\bigr]\Bigr),
\end{aligned}
\end{equation}
with adaptive scaling \(\beta^{\text{Margin}}=L(x,\tau,\lambda,\mu)/f(x,\tau_\star)>1\).  
The Lagrangian \(L\) smooths the transition into the feasible region, eliminating gradient discontinuities. This loss simultaneously attracts the policy toward the current best feasible solution and repels it from all infeasible ones.

\paragraph{Primal Refinement Loss \(\mathcal L^{\text{Primal}}\).}  
Activated when \(|\mathcal T^{\!T}|\ge 2\). We refine the policy within the feasible region by pairing \(\tau_\star\) with every other feasible solutions \(\tau\in\mathcal T^{\!T}\setminus\{\tau_\star\}\):
\begin{equation}
\begin{aligned}
\mathcal{L}^{\text{Primal}}&(\theta)=
  -\mathbb{E}_{x\sim\mathcal{X}}
\frac{1}{|\mathcal{T}^{\!T}|-1}\sum_{\tau\in\mathcal{T}^{\!T}\setminus\{\tau_\star\}}
\log\sigma\!\Bigl(\\
  &\beta^{\text{Primal}}\bigl[\log\pi_\theta(\tau_\star\!\mid\!x)-\log\pi_\theta(\tau\!\mid\!x)\bigr]\Bigr),
\end{aligned}
\end{equation}
scaled by \(\beta^{\text{Primal}}=f(x,\tau)/f(x,\tau_\star)\ge 1\).
Empirically, we observe that $\beta^{\text{Margin}}>\beta^{\text{Primal}}$ in most instances; the larger gradient on the constraint term implicitly prioritizes feasibility. If $\mid \mathcal{T}^{\!T}\mid\leq 1$, we simply set $\mathcal{L}^{\text{Primal}}=0$.

\paragraph{Overall Training Objective.}  
The policy network is optimized with the composite loss,
\begin{equation}
\mathcal{L}(\theta)=\mathcal{L}^{\text{Dual}}(\theta)+\mathcal{L}^{\text{Margin}}(\theta)+\mathcal{L}^{\text{Primal}}(\theta),
\label{eq:overallloss}
\end{equation}
where the active terms are automatically selected from cold start to high performance, ensuring stable end-to-end training across all stages.

\subsection{Light Warm-Start Fine-Tuning}

\begin{algorithm}[t]
\caption{UCPO Training Protocol.}
\label{alg1}
\begin{algorithmic}[1]
\Require
    Epochs $E_{\text{ft}}$, batch size $B$, data distribution $S$, NCO pretrained checkpoint $\theta_0$.
\State \textbf{Initialize} $\theta\gets\theta_0$
\For{epoch $e = 1$ \textbf{to} $E_{\text{ft}}$}
    \For{batch $b = 1$ \textbf{to} $\lceil E_{\text{ft}}/B \rceil$}
        \State $\bigl\{\langle x_i,\mathcal{G}_i,\mathcal{H}_i\rangle\bigr\}_{i=1}^{B}\gets \text{DataGenerator}(S)$
        \For{each $x$ in mini-batch}
            \State Sample solutions
                $\tau_{(1\dots N)} \sim \text{NCO}\bigl(\langle x_i,\mathcal{G}_i,\mathcal{H}_i\rangle\bigr)$
            \State Sort and split solutions
            \State\quad
                $\mathcal{T}^{T}\gets\{\tau\in\mathcal{T}_x \mid I(x,\tau)=0\}$
            \State\quad
                $\mathcal{T}^{F}\gets\{\tau\in\mathcal{T}_x \mid I(x,\tau)=1\}$
            \State Compute $\mathcal{L}^{\text{Dual}}(\theta), \mathcal{L}^{\text{Margin}}(\theta),\mathcal{L}^{\text{Primal}}(\theta)$
            \State $\mathcal{L}(\theta)=\mathcal{L}^{\text{Dual}}(\theta)
                +\mathcal{L}^{\text{Margin}}(\theta)
                +\mathcal{L}^{\text{Primal}}(\theta)$
            \State Update parameters
                $\theta\gets\theta-\alpha\nabla_\theta\mathcal{L}(\theta)$
        \EndFor
    \EndFor
\EndFor
\end{algorithmic}
\end{algorithm}

The training procedure of UCPO, as formalized in Algorithm \ref{alg1}, is designed to seamlessly integrate with existing NCO frameworks. It achieves full architectural compatibility by neither modifying the network topology nor introducing additional learnable parameters. This design allows any pre-trained checkpoint, especially those already fine-tuned for constrained settings, to serve as a high-quality initializer.

Given a pre-trained policy \(\pi_{\theta_0}\), UCPO performs fine-tuning for a limited number of epochs \(E_{\text{ft}} \ll E_{\text{base}}\), where \(E_{\text{base}}\) denotes the original training epochs. The fine-tuning objective minimizes the composite loss defined in Eq. (\ref{eq:overallloss}) through gradient updates. This process simultaneously enhances constraint satisfaction and solution quality.

The warm-start protocol achieves two key advantages:
\begin{itemize}
    \item \textbf{Few-Overhead Transfer}: By preserving architectural invariance, UCPO ensures that the computational complexity of the pre-trained model remains unchanged. Specifically, the memory footprint and inference latency of the fine-tuned model are equivalent to those of the pre-trained model.
    \item \textbf{Rapid Convergence}: Empirically, UCPO reaches stable performance within \(E_{\text{ft}} = \lceil 0.01E_{\text{base}} \rceil\) to \(\lceil 0.05E_{\text{base}} \rceil\) epochs. This efficiency enables UCPO to achieve significant performance improvements with less than 5\% of the original training budget.
\end{itemize}

\section{Experiments}

\begin{table*}[!t]
  \centering
  \small
  \renewcommand{\arraystretch}{1.0}
  \setlength{\tabcolsep}{1.4pt}
  \newcommand{\gr}[1]{\cellcolor{black!10}#1}
  \newcommand{\B}[1]{\textbf{#1}}
  \begin{tabular}{@{} l
    r r r r r r r r r
    r r r r r r r r r @{}}
    \toprule
    \multirow{3}{*}{Method} &
    \multicolumn{9}{c}{\textbf{TSPTW-50}} &
    \multicolumn{9}{c}{\textbf{TSPTW-100}} \\
    \cmidrule(lr){2-10} \cmidrule(lr){11-19}
    & \multicolumn{3}{c}{Easy} & \multicolumn{3}{c}{Medium} & \multicolumn{3}{c}{Hard}
    & \multicolumn{3}{c}{Easy} & \multicolumn{3}{c}{Medium} & \multicolumn{3}{c}{Hard} \\
    \cmidrule(lr){2-4} \cmidrule(lr){5-7} \cmidrule(lr){8-10}
    \cmidrule(lr){11-13} \cmidrule(lr){14-16} \cmidrule(lr){17-19}
    &   Inst.\% & Obj. & Gap\% &   Inst.\% & Obj. & Gap\% &   Inst.\% & Obj. & Gap\%
    &   Inst.\% & Obj. & Gap\% &   Inst.\% & Obj. & Gap\% &   Inst.\% & Obj. & Gap\% \\
    \midrule
    LKH3               & 0.00 & 7.31 & 0.00 & 0.00 & 13.02 & 0.00 & 0.12 & 25.61 & 0.00 & 0.00 & 10.21 & 0.00 & 0.00 & 18.74 & 0.00 & 0.07 & 51.24 & 0.00 \\
    \midrule
    POMO               & 0.00 & 7.54 & 3.15 & 3.77 & 13.68 & 5.07 & 35.25 & 26.22 & 2.38 & 0.00 & 10.83 & 6.07 & 0.12 & 20.78 & 10.89 & 100.00 & — & — \\
    \gr{UCPO (POMO)}   & 0.00 & 7.41 & 1.30 & 0.08 & 13.26 & 1.86 & 1.32 & 25.75 & 0.55 & 0.00 & 10.54 & 3.25 & 0.01 & 19.84 & 5.89 & 100.00 & — & — \\
    \gr{UCPO* (POMO)} & 0.00 & \B{7.36} & \B{0.64} & 0.15 & 13.19 & 1.31 & 0.13 & 25.61 & 0.01 & 0.00 & 10.58 & 3.66 & 0.00 & 19.93 & 6.35 & \B{0.87} & 51.25 & 0.02 \\
    \midrule
    PIP                & 0.00 & 7.50 & 2.60 & 0.90 & 13.40 & 2.92 & 2.67 & 25.66 & 0.20 & 0.00 & 10.57 & 3.53 & 0.19 & 19.61 & 4.64 & 16.27 & 51.42 & 0.35 \\
    \gr{UCPO (PIP)}    & 0.00 & 7.39 & 1.11 & \B{0.04} & \B{13.14} & \B{0.89} & \B{0.76} & \B{25.61} & \B{0.00}
                      & 0.00 & \B{10.41} & \B{1.94} & \B{0.00} & \B{19.17} & \B{2.27} & 5.01 & 51.28 & 0.07 \\
    \midrule
    PIP-D              & 0.00 & 7.49 & 2.46 & 0.65 & 13.45 & 3.30 & 3.07 & 25.69 & 0.31 & 0.00 & 10.66 & 4.41 & 0.03 & 19.79 & 5.60 & 6.48 & 51.39 & 0.29 \\
    \gr{UCPO (PIP-D)}  & 0.00 & 7.42 & 1.49 & 0.05 & 13.18 & 1.23 & 1.07 & 25.61 & 0.00 & 0.00 & 10.53 & 3.11 & 0.01 & 19.32 & 3.08 & 2.99 & \B{51.24} & \B{0.00} \\
    \bottomrule
  \end{tabular}
  \caption{Performance on TSPTW-50 and TSPTW-100. Bars indicate no feasible solution. UCPO* represents UCPO with cold-start training.}
  \label{tab:tsptw}
\end{table*}

\begin{table}[!t]
  \centering
  \small
  \renewcommand{\arraystretch}{1.0}
  \setlength{\tabcolsep}{3pt} 
  \newcommand{\hdr}[1]{\multicolumn{1}{c}{#1}}
  \newcommand{\gr}[1]{\cellcolor{black!10}#1}
  \newcommand{\B}[1]{\textbf{#1}}
  \begin{tabular}{l r r r r r r}
    \toprule
    \multirow{2}{*}{Method} & \multicolumn{3}{c}{Medium (TSPDL-50)} & \multicolumn{3}{c}{Hard (TSPDL-50)} \\
    \cmidrule(lr){2-4} \cmidrule(lr){5-7}
    & \hdr{Inst.\%} & \hdr{Obj.} & \hdr{Gap\%} & \hdr{Inst.\%} & \hdr{Obj.} & \hdr{Gap\%} \\
    \midrule
    LKH3              & 0.00 & 10.87 & 0.00 & 0.00 & 13.30 & 0.00 \\
    \midrule
    POMO              & 12.52 & 10.98 & 1.01 & 29.25 & 13.85 & 4.10 \\
    \gr{UCPO (POMO)}  & 4.00 & 11.04 & 1.54 & 0.01 & \B{13.48} & \B{1.35} \\
    \gr{UCPO*(POMO)}  & 0.07 & 11.09 & 2.03 & \B{0.00} & 13.54 & 1.83 \\
    \midrule
    PIP               & 0.43 & 11.22 & 3.22 & 2.10 & 13.66 & 2.71 \\
    \gr{UCPO (PIP)}   & 0.02 & \B{11.06} & \B{1.73} & 0.01 & 13.48 & 1.36 \\
    \midrule
    PIP-D             & 0.37 & 11.26 & 3.59 & 0.82 & 13.80 & 3.76 \\
    \gr{UCPO (PIP-D)} & \B{0.01} & 11.10 & 2.15 & 0.04 & 13.56 & 1.92 \\
    \midrule \addlinespace[-1ex] \midrule
    \multirow{2}{*}{Method} & \multicolumn{3}{c}{Medium (TSPDL-100)} & \multicolumn{3}{c}{Hard (TSPDL-100)} \\
    \cmidrule(lr){2-4} \cmidrule(lr){5-7}
    & \hdr{Inst.\%} & \hdr{Obj.} & \hdr{Gap\%} & \hdr{Inst.\%} & \hdr{Obj.} & \hdr{Gap\%} \\
    \midrule
    LKH3              & 0.00 & 16.39 & 0.00 & 0.00 & 20.70 & 0.00 \\
    POMO              & 32.16 & 17.11 & 4.39 & 99.85 & \B{20.95} & \B{1.21} \\
    \gr{UCPO (POMO)}  & \B{0.02} & 17.29 & 5.49 & 18.84 & 21.40 & 3.38 \\
    \gr{UCPO*(POMO)}  & 0.03 & 17.27 & 5.38 & \B{0.01} & 22.37 & 8.06 \\
    \midrule
    PIP               & 0.38 & 17.71 & 8.05 & 20.66 & 22.30 & 7.73 \\
    \gr{UCPO (PIP)}   & \B{0.02} & \B{17.19} & \B{4.86} & 0.04 & 22.40 & 8.22 \\
    \midrule
    PIP-D             & 0.23 & 17.84 & 8.85 & 7.91 & 22.84 & 10.34 \\
    \gr{UCPO (PIP-D)} & \B{0.02} & 17.44 & 6.42 & \B{0.01} & 22.42 & 8.30 \\
    \bottomrule
  \end{tabular}
  \caption{Performance on TSPDL-50 and TSPDL-100.}
  \label{tab:tspdl}
\end{table}

\begin{table}[t]
  \centering
  \small
  \renewcommand{\arraystretch}{1.0}
  \setlength{\tabcolsep}{2.8pt}
  \newcommand{\gr}[1]{\cellcolor{black!10}#1}
  \newcommand{\B}[1]{\textbf{#1}}
  \begin{tabular}{@{} l r@{\,}r r@{\,}r r @{}}
    \toprule
    \multirow{2}{*}{Method} &
    \multicolumn{2}{c}{CVRPTW50} &
    \multicolumn{2}{c}{CVRPTWLV50} \\
    \cmidrule(lr){2-3} \cmidrule(lr){4-5}
    & Inst.\ (\%) & Obj. & Inst.\ (\%) & Obj. \\
    \midrule
    POMO        & 0.20 & 14.13 & 3.26 & 14.20 \\
    \gr{UCPO (POMO)}  & \B{0.09} & 13.58 & 0.98 & 13.61 \\
    \gr{UCPO* (POMO)} & 0.75 & \B{13.50} & 2.20 & \B{13.56} \\
    \midrule
    SLIM        & 0.15 & 14.10 & 2.74 & 14.15 \\
    \gr{UCPO (SLIM)}  & 0.11 & 13.51 & \B{0.85} & 13.59 \\
    \bottomrule
  \end{tabular}
  \caption{Performance on CVRPTW50 and CVRPTWLV50.}
  \label{tab:cvrptw}
\end{table}

\begin{table}[t]
  \centering
  \small
  \renewcommand{\arraystretch}{1.0}
  \setlength{\tabcolsep}{3.4pt}
  \newcommand{\B}[1]{\textbf{#1}}
  \begin{tabular}{@{} l r@{\,}r r@{\,}r @{}}
    \toprule
    \multirow{2}{*}{Method} &
    \multicolumn{2}{c}{Medium} &
    \multicolumn{2}{c}{Hard} \\
    \cmidrule(lr){2-3} \cmidrule(lr){4-5}
    & Inst.\ (\%) & Gap\ (\%) & Inst.\ (\%) & Gap\ (\%) \\
    \midrule
    LKH3                & 0.00 & 0.00 & 0.12 & 0.00 \\
    \midrule
    UCPO-PIP ($\lambda$=0.5) & \B{0.01} & 0.95 & \B{0.24} & 0.00 \\
    UCPO-PIP ($\lambda$=1)   & 0.04 & \B{0.89} & 0.76 & 0.00 \\
    UCPO-PIP ($\lambda$=2)   & 0.03 & 0.94 & 0.78 & \B{-0.01} \\
    \midrule
    UCPO-POMO ($\lambda$=0.5)& \B{0.03} & 1.97 & 1.42 & 0.61 \\
    UCPO-POMO ($\lambda$=1)  & 0.08 & \B{1.86} & \B{1.32} & \B{0.55} \\
    UCPO-POMO ($\lambda$=2)  & 0.09 & 1.87 & 1.77 & 0.61 \\
    \bottomrule
  \end{tabular}
  \caption{Ablation on TSPTW-50 with different $\lambda$ values.}
  \label{tab:abl}
\end{table}

\subsection{Experimental Setups}

\subsubsection{Datasets.} Experiments target four canonical constraints: time windows, draft limits, vehicle capacity, and fleet size. Four benchmark tasks are examined accordingly: TSPTW, TSPDL, CVRPTW, and CVRPTWLV. For TSPTW and TSPDL, we adopt the generation protocol proposed by PIP and create instances at three difficulty levels—Easy, Medium, and Hard—based on constraint-satisfaction hardness. CVRPTW and CVRPTWLV follow the setting introduced by JAMPR \cite{falkner2020learning} and are generated uniformly at the Hard level. All datasets are fixed at 50 and 100 nodes, with detailed generative procedures provided in the Appendix C. Each validation set comprises exactly 10,000 instances.


\subsubsection{Baselines.} UCPO is evaluated against three representative algorithmic families: 1) Problem-specific heuristics, exemplified by LKH3 \cite{helsgaun2017extension}, which are meticulously tailored to each VRP variant. 2) General-purpose RL-NCO, represented by POMO \cite{kwon2020pomo} and SLIM \cite{corsini2024self}, that employs reinforcement learning to deliver universal solvers. 3) Constraint-specialized models, including PIP \cite{bi2024learning} and its distilled variant PIP-D.

\subsubsection{Benchmark Setup and Base Models.} UCPO is warm-started into three backbone models: POMO, PIP, and PIP-D, with their architectures detailed in the Appendix B. All hyper-parameters are kept identical to the original publications, and the publicly released best checkpoints are used without modification. Training data distributions, network architectures, and optimizer configurations remain exactly the same. For Lagrangian-based models, the Lagrange multiplier is fixed to 1, consistent with the reported values. During warm-start, POMO and PIP are fine-tuned for up to 500 epochs; PIP-D is trained for 100 epochs. These epoch counts are jointly determined by convergence curves and training-time budgets. Note that the original checkpoints were obtained after 10,000 epochs, so warm-start training amounts to only 1\%–5\% of the total previous computation. To provide a comprehensive evaluation, UCPO is additionally cold-started on POMO without pre-trained checkpoint. All training and inference are executed on NVIDIA GeForce GTX 1080 Ti GPUs and Intel(R) Xeon(R) CPU E5-2640 v4 @ 2.40GHz CPUs.


\subsubsection{Evaluation.} Inference follows a unified protocol: 1) Sampling-based decoding: the number of samples equals the instance size; 8× data augmentation is enabled. 2) Metrics include the infeasible rate on the validation set, the average objective value of \textbf{feasible solutions} and the average optimality gap with respect to LKH3. An instance is regarded as feasible if at least one sampled solution satisfies all constraints; the solution with the smallest objective among the feasible ones is returned. 3) Inference time and memory usage are not reported, as they remain consistent with the backbone models.

\subsection{Main Results}

The performance of UCPO on various benchmark tasks is summarized in Tables \ref{tab:tsptw}--\ref{tab:cvrptw}. After fine-tuning with UCPO, the NCO models based on different backbones, including POMO, PIP, and PIP-D, consistently achieve substantial improvements in both feasibility rate and objective value across diverse test sets.

As illustrated in Table \ref{tab:tsptw}, UCPO achieves remarkable performance on TSPTW instances. First, it significantly reduces infeasibility rates. On the Easy and Medium sets, UCPO attains near-zero infeasibility rates (all below 0.15\%), significantly outperforming PIP (0.90\%) and PIP-D (0.65\%). On the Hard set, UCPO* (POMO) reduces POMO's infeasibility rate from 100\% to 0.87\%. Second, UCPO substantially improves solution quality. On TSPTW-50-Easy, UCPO* (POMO) reduces the optimality gap from 3.15\% to 0.64\%, achieving superior performance compared to all other methods. On TSPTW-100-Easy, UCPO (PIP) lowers the gap to 1.94\%, outperforming PIP’s 3.53\%. Third, UCPO exhibits strong generalization and convergence across different backbones. It consistently enhances performance across various policy networks (POMO, PIP, PIP-D), reducing the performance gap between baseline models. For example, UCPO (PIP) ranks first on four out of six subsets. Fourth, UCPO* achieves substantial infeasibility reduction (e.g., from 16.27\% to 0.87\% on TSPTW-100-Hard), although it requires significantly more training time to converge.

Tables \ref{tab:tspdl} and \ref{tab:cvrptw} further demonstrate UCPO’s consistent improvements in feasibility and objective values over baseline methods. On TSPDL-100-Hard, UCPO* (POMO) reduces infeasibility from 99.85\% to 0.01\%, achieving better trade-offs. Even in cold-start mode on CVRPTW, UCPO* achieves the best objective values (13.50 vs. 14.13), showcasing its strong generalization capability under hard constraints.

\begin{table}[t]
  \centering
  \small
  \renewcommand{\arraystretch}{1.0}
  \setlength{\tabcolsep}{2pt}
  \newcommand{\B}[1]{\textbf{#1}}
  \begin{tabular}{@{} ccc r@{\,}r r@{\,}r r@{\,}r @{}}
    \toprule
    \multicolumn{3}{c}{Loss} & \multicolumn{2}{c}{Medium} & \multicolumn{2}{c}{Hard} \\
    \cmidrule(lr){1-3} \cmidrule(lr){4-5} \cmidrule(lr){6-7}
    Feasibility & Primal & Dual & Inst.\ (\%) & Gap\ (\%) & Inst.\ (\%) & Gap\ (\%) \\
    \midrule
    \checkmark &      &      & 0.29 & 1.87 & 1.74 & 0.64 \\
         & \checkmark &      & 0.30 & \B{1.84} & 1.91 & 0.62 \\
         &      & \checkmark & 100.00 & — & 100.00 & — \\
    \checkmark & \checkmark &      & 0.21 & 1.87 & 1.94 & 0.59 \\
    \checkmark &      & \checkmark & 0.14 & 1.90 & 1.53 & 0.62 \\
         & \checkmark & \checkmark & 0.19 & 1.90 & 2.19 & 0.62 \\
    \checkmark & \checkmark & \checkmark & \B{0.08} & 1.86 & \B{1.32} & \B{0.55} \\
    \midrule
    \checkmark & \checkmark &  & 100.00 & — & 100.00 & — \\
    \checkmark & \checkmark & \checkmark & \B{0.08} & \B{1.31} & \B{1.61} & \B{0.01} \\
    \bottomrule
  \end{tabular}
  \caption{Ablation of loss terms on TSPTW-50. The upper pane presents results obtained with warm-started models; the lower pane corresponds to models trained from scratch.}
  \label{tab:loss}
\end{table}

\begin{figure}[t]
\centering
\begin{subfigure}[t]{0.45\textwidth}
  \centering
  \includegraphics[width=\linewidth]{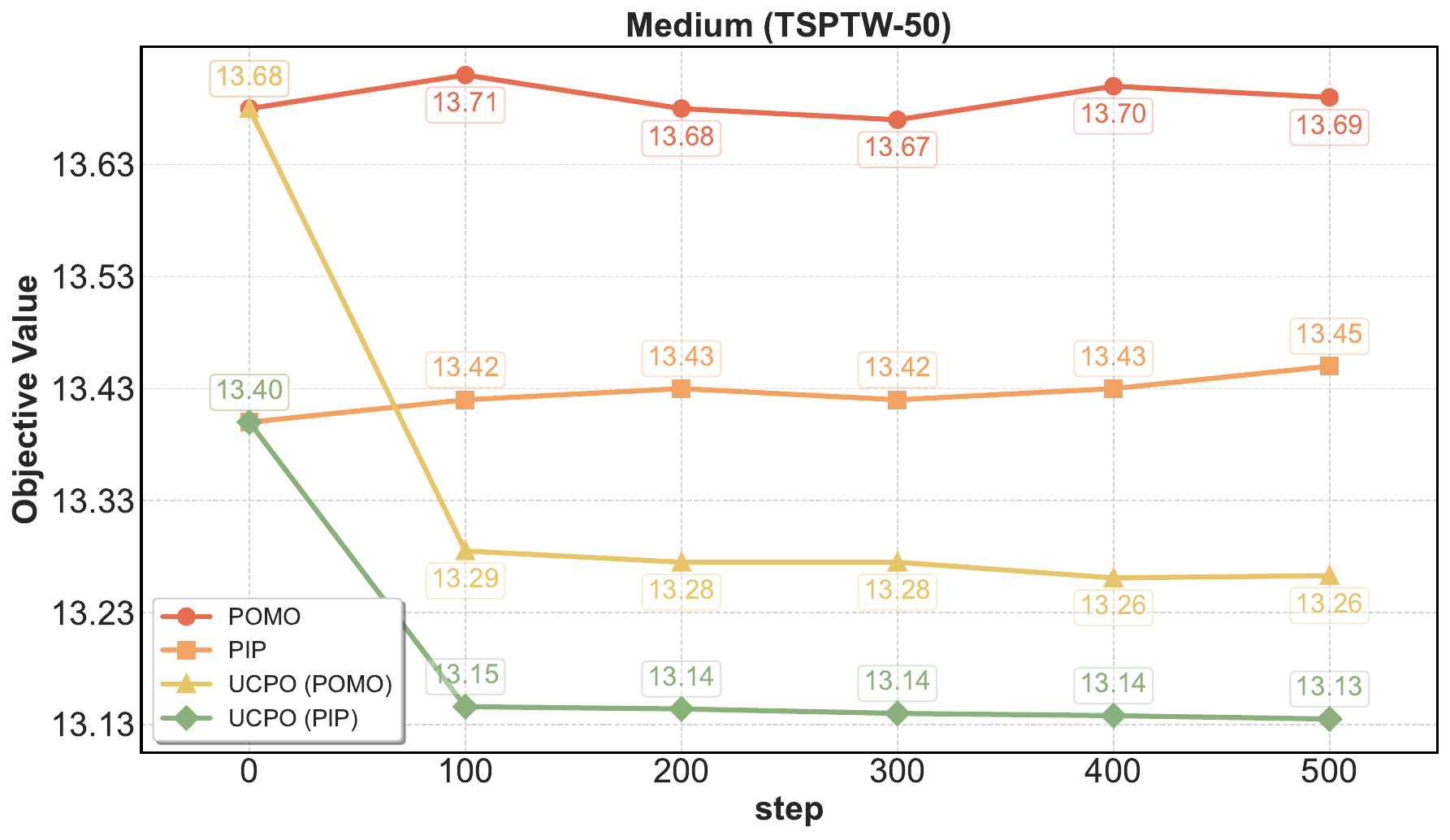}
  \caption{Medium}
\end{subfigure}%
\hfill
\begin{subfigure}[t]{0.45\textwidth}
  \centering
  \includegraphics[width=\linewidth]{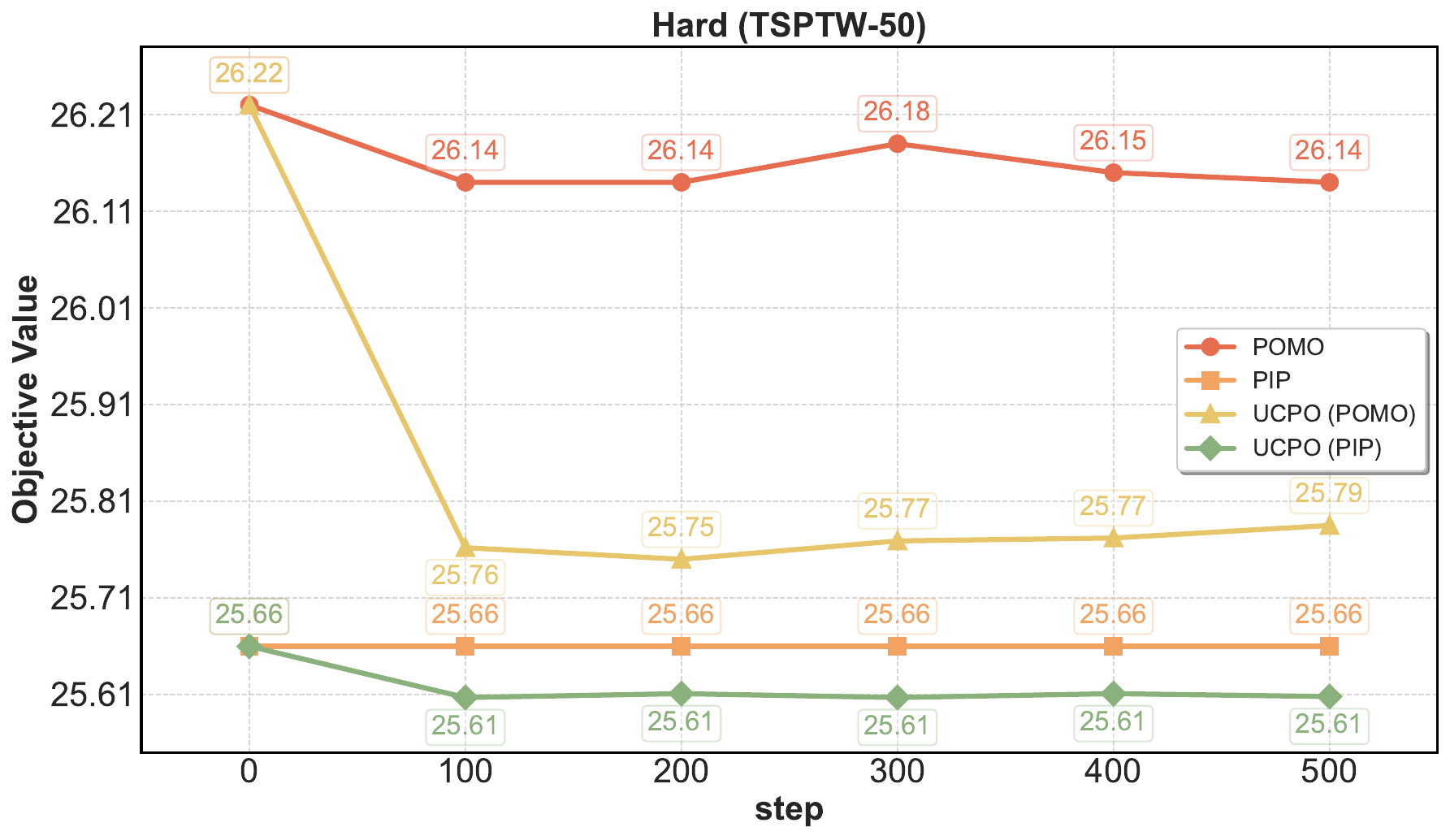}
  \caption{Hard}
\end{subfigure}
\caption{Post-training values across two difficulty levels (Medium, Hard).}
\label{fig:post_train_values}
\end{figure}

\subsection{Sensitivity Analysis}

To verify the robustness of UCPO with respect to the Lagrange multiplier $\lambda$, we conduct experiments on 50-node TSPTW instances on both Medium and Hard levels. Using POMO and PIP as base models, we train and evaluate with $\lambda \in \{0.5, 1, 2\}$.

As shown in Table \ref{tab:abl}, the maximal deviations in infeasibility rate and objective gap across different $\lambda$ are merely 0.54\% and 0.11\%, respectively. Moreover, neither metric exhibits a monotonic trend with respect to $\lambda$. The observed variations can bee attributable to training stochasticity. These results indicate that UCPO exhibits minimal sensitivity to its sole hyperparameter $\lambda$, underscoring its exceptional parameter robustness. This characteristic is particularly significant in practical applications, as it eliminates the need for meticulous hyperparameter tuning and enhances the general applicability of UCPO across diverse constrained optimization scenarios.

\subsection{Ablation Studies}

We conduct systematic ablations to assess the contributions of the three components of our universal constrained loss function and the impact of warm-start training epochs.

As shown in Table \ref{tab:loss}, experiments on 50-node TSPTW instances (Medium and Hard difficulty levels) yield two key insights. 1) Both the \emph{Feasibility Margin Loss} and \emph{Primal Refinement Loss} are essential for achieving optimal performance. Removing either component results in a measurable decline in feasibility or solution quality, whereas combining these two losses consistently delivers superior outcomes. 2) The \emph{Dual Exploration Loss}, although inactive during warm-start training (due to the high feasibility of generated trajectories), proves indispensable in cold-start scenarios, enabling rapid convergence to feasible solutions.

Figure \ref{fig:post_train_values} further illustrates the efficiency of our warm-start approach. After further training from converged POMO or PIP checkpoints yields minimal improvements, UCPO achieves significant performance gains after just 100 warm-start steps, representing merely 1\% of the original training budget. Additional ablation results and analyses are provided in the Appendix D.

\section{Conclusion}

We introduce Universal Constrained Preference Optimization (UCPO), a novel framework that addresses the feasibility–performance trade-off in neural combinatorial optimization (NCO) with complex constraints. UCPO is designed as a plug-and-play module that can be integrated with any existing NCO model without requiring architectural changes or manual tuning of Lagrange multipliers. By leveraging a warm-start fine-tuning procedure, UCPO achieves near-100\% feasibility and near-optimal solutions on complex constrained tasks with just 1\%–5\% of the original training budget.

For future research, we propose three directions: 1) developing mechanisms for UCPO to handle dynamic constraints in real-time applications, 2) extending UCPO with meta-learning capabilities to enable rapid adaptation to unseen constraints, and 3) decoupling the constraint module to allow a single base model to generalize across diverse constraint settings through module swapping.

\bigskip

\bibliography{aaai2026}

\appendix

\begin{center}
  \Large\bfseries Appendix
\end{center}

\section{A. Problems Details}

\subsection{A.1 The Traveling Salesman Problem with Time Windows (TSPTW)}

TSPTW has extensive applications across various domains such as logistics distribution, vehicle routing, production scheduling, and urban service systems. For instance, in logistics operations, companies often need to perform pickups or deliveries within specific time windows designated by customers, which represents a typical TSPTW scenario. Compared with the classical TSP, TSPTW introduces an additional requirement that each city must be visited within its specified time window.

\subsubsection{Problem Description.}  
Given a complete directed graph \( G = (V, A) \), the node set is defined as \( V = \{0, 1, \dots, n\} \), where node \( 0 \) represents the depot (the start and end point), and the remaining \( n \) nodes correspond to customers. The arc set \( A \subseteq V \times V \) denotes the feasible paths between any two nodes. The travel cost \( c_{ij} \) is defined as the distance or travel time associated with the arc \( (i, j) \in A \). The time window \( [e_i, l_i] \) indicates the interval during which service must commence at customer \( i \). The service time \( s_i \) represents the dwell time at node \( i \), which is set to \( s_i = 0 \) in this paper.

\subsubsection{Decision Variables.}
\begin{itemize}
    \item \( x_{ij} \in \{0, 1\} \): Takes the value 1 if the tour proceeds directly from node \( i \) to node \( j \), and 0 otherwise.
    \item \( t_i \ge 0 \): Represents the arrival time at node \( i \).
\end{itemize}

\subsubsection{Mixed-Integer Linear Programming Model.}

\begin{equation}
\min\; Z = \sum_{i \in V} \sum_{j \neq i} c_{ij} x_{ij}
\end{equation}
\begin{equation}
\sum_{j \neq i} x_{ij} = 1 \quad \forall i \in V
\end{equation}
\begin{equation}
\sum_{i \neq j} x_{ij} = 1 \quad \forall j \in V
\end{equation}
\begin{equation}
e_i \le t_i \le l_i \quad \forall i \in V
\end{equation}
\begin{equation}
t_j \ge t_i + s_i + c_{ij} x_{ij} - M (1 - x_{ij}) \quad \forall i, j \in V, \; i \neq j
\end{equation}
\begin{equation}
u_i - u_j + n \, x_{ij} \le n - 1 \quad \forall i, j \in V \setminus \{0\}, \; i \neq j
\end{equation}
\begin{equation}
x_{ij} \in \{0, 1\} \quad \forall i,j\in V,i \neq j
\end{equation}
\begin{equation}
t_i \ge 0 \quad \forall i \in V
\end{equation}

The objective function aims to minimize the total travel distance. Constraints (2) and (3) are out-degree and in-degree constraints that ensure each node is visited exactly once. Constraints (4) and (5) are time window and time continuity constraints that enforce service to start within the specified interval and maintain the logical sequence of the tour. Constraints (6) are the MTZ constraints, which eliminate subtours and guarantee the formation of a single Hamiltonian circuit.

\subsection{A.2 The Traveling Salesman Problem with Draft Limit (TSPDL)}

TSPDL is widely applied in maritime logistics, inland waterway transportation, and feeder shipping, where vessel navigation is restricted by port or channel-specific draft limits. In practice, a shipping line must design a port-calling sequence that minimizes sailing cost (or time) while ensuring the vessel’s arrival draft never exceeds the maximum permissible draft at any visited port. Compared with the classical TSP, TSPDL additionally imposes a draft feasibility constraint that is dynamically coupled with the residual cargo on board.

\subsubsection{Problem Description.} Consider a complete directed graph \( G = (V, A) \), where the node set \( V = \{0, 1, \dots, n\} \) consists of the loading port (depot) denoted as node 0 and the remaining \( n \) nodes representing discharge ports. The arc set \( A \subseteq V \times V \) represents the navigable legs between any two ports. The sailing cost \( c_{ij} \) is defined as the distance, fuel consumption, or travel time for the arc \( (i, j) \in A \). Each port \( i \) has a maximum draft limit \( D_i > 0 \), which specifies the maximum permissible draft (in metres) allowed at that port. The cargo demand \( d_i \ge 0 \) indicates the quantity (in tonnes, TEU, or draft-equivalent units) to be discharged at port \( i \), with the total load \( Q = \sum_{i \in V \setminus \{0\}} d_i \). The initial draft at the loading port (node 0) is \( D_0 \ge Q \), meaning the vessel is fully laden when departing from port 0.

\subsubsection{Decision Variables.}
\begin{itemize}
    \item \( x_{ij} \in \{0, 1\} \): Equals 1 if the vessel sails directly from port \( i \) to port \( j \), and 0 otherwise.
    \item \( y_i \ge 0 \): Residual cargo remaining on board when departing from port \( i \).
    \item \( z_i \ge 0 \): Arrival draft at port \( i \), measured before any cargo operations.
\end{itemize}

\subsubsection{Mixed-Integer Linear Programming Model.}

\begin{equation}
\min\; Z = \sum_{i\in V}\sum_{j\neq i} c_{ij}x_{ij}
\end{equation}
\begin{equation}
\sum_{j\neq i} x_{ij} = 1 \quad \forall i\in V
\end{equation}
\begin{equation}
\sum_{i\neq j} x_{ij} = 1 \quad \forall j\in V
\end{equation}
\begin{equation}
y_0 = Q
\end{equation}
\begin{equation}
y_j \le y_i - d_j + M(1 - x_{ij}) \quad \forall i,j\in V,\, j\neq i
\end{equation}
\begin{equation}
z_i = y_i \quad \forall i\in V
\end{equation}
\begin{equation}
z_i \le D_i \quad \forall i\in V
\end{equation}
\begin{equation}
u_i - u_j + n x_{ij} \le n - 1 \quad \forall i,j\in V\!\setminus\!\{0\},\, i\neq j
\end{equation}
\begin{equation}
x_{ij} \in \{0,1\} \quad \forall i,j\in V,i\neq j
\end{equation}
\begin{equation}
y_i,\, z_i \ge 0 \quad \forall i\in V
\end{equation}

The objective function aims to minimize the total sailing cost. Constraints (10) and (11) are out-degree and in-degree constraints that ensure each port is visited exactly once. Constraints (12) and (13) manage cargo balance, tracking the residual cargo after each discharge. Constraints (14) and (15) enforce draft feasibility at every port: the draft–cargo identity \( z_i = y_i \) and the draft limit \( z_i \le D_i \). Constraints (16) are the MTZ subtour elimination constraints, which guarantee the formation of a single Hamiltonian circuit.

\subsection{A.3 Capacitated Vehicle Routing Problem with Time Windows (CVRPTW)}

CVRPTW is a cornerstone model in freight distribution and urban logistics, especially for last-mile delivery. In practice, a fleet of identical vehicles, stationed at a single depot, must serve geographically dispersed customers. Each customer has known demands and strict service time windows, and vehicle capacity must not be exceeded. Unlike the classical VRP, CVRPTW incorporates both capacity constraints and time-window requirements for each customer.

\subsubsection{Problem Description.} 
Consider a complete directed graph \( G = (V, A) \), where the node set \( V = \{0, 1, \dots, n\} \) includes the depot (node 0, the start and end of all routes) and customer nodes \( V_{\text{c}} = \{1, \dots, n\} \). The arc set \( A \subseteq V \times V \) represents all feasible connections between nodes. The travel cost \( c_{ij} \) is defined as the distance, travel time, or fuel consumption on arc \( (i, j) \in A \). Each customer \( i \in V_{\text{c}} \) has a demand \( q_i > 0 \), while the depot has \( q_0 = 0 \). The vehicle capacity is \( Q > 0 \), and all vehicles in the fleet have the same capacity. Each node \( i \) has a time window \( [e_i, l_i] \), during which service must commence. The service time \( s_i \ge 0 \) represents the dwell time at node \( i \), with \( s_i = 0 \) at the depot. The fleet size \( K \) is assumed to be \emph{unlimited}, but the model implicitly determines the minimum number of vehicles required.

\subsubsection{Decision Variables.}
\begin{itemize}
    \item \( x_{ij}^k \in \{0, 1\} \): Equals 1 if vehicle \( k \) travels directly from node \( i \) to node \( j \), and 0 otherwise.
    \item \( t_i^k \ge 0 \): Arrival time of vehicle \( k \) at node \( i \).
    \item \( u_i^k \ge 0 \): Cumulative load on vehicle \( k \) when departing from node \( i \).
\end{itemize}

\subsubsection{Mixed-Integer Linear Programming Model.}

\begin{equation}
\min\; Z = \sum_{k=1}^{K} \sum_{i\in V} \sum_{j\neq i} c_{ij}\,x_{ij}^{k}
\end{equation}
\begin{equation}
\sum_{k=1}^{K} \sum_{j\neq i} x_{ij}^{k} = 1 \quad \forall i\in V_{\!c}
\end{equation}
\begin{equation}
\sum_{j\neq i} x_{ij}^{k} - \sum_{j\neq i} x_{ji}^{k} = 0 \quad \forall i\in V,\;\forall k=1,\dots,K
\end{equation}
\begin{equation}
\sum_{j\in V_{\!c}} x_{0j}^{k} \le 1 \quad \forall k=1,\dots,K
\end{equation}
\begin{equation}
\sum_{i\in V_{\!c}} x_{i0}^{k} \le 1 \quad \forall k=1,\dots,K
\end{equation}
\begin{equation}
u_{j}^{k} \ge u_{i}^{k} + q_{j} - Q(1 - x_{ij}^{k}) \quad \forall i,j\in V,\, j\neq i,\, \forall k
\end{equation}
\begin{equation}
u_{i}^{k} \le Q \quad \forall i\in V,\;\forall k=1,\dots,K
\end{equation}
\begin{equation}
e_{i} \le t_{i}^{k} \le l_{i} \quad \forall i\in V,\;\forall k=1,\dots,K
\end{equation}
\begin{equation}
t_{j}^{k} \ge t_{i}^{k} + s_{i} + c_{ij} - M(1 - x_{ij}^{k}) \quad \forall i,j\in V,\, j\neq i,\, \forall k
\end{equation}
\begin{equation}
x_{ij}^{k} \in \{0,1\} \quad \forall i\neq j,\;\forall k=1,\dots,K
\end{equation}
\begin{equation}
t_{i}^{k},\; u_{i}^{k} \ge 0 \quad \forall i\in V,\;\forall k=1,\dots,K
\end{equation}

The objective function is to minimize the total travel cost across all routes. Constraints (20) and (21) ensure that each customer is visited exactly once and that the routes form contiguous paths. Constraints (22) and (23) limit the number of vehicles leaving and returning to the depot. Constraints (24) and (25) enforce vehicle capacity limits throughout each tour. Constraints (26) and (27) ensure that service starts within the prescribed time windows and maintain the logical sequence of the tour.


\subsection{A.4 Capacitated Vehicle Routing Problem with Time Windows and Limited Vehicle (CVRPTWLV)}

CVRPTWLV is a direct extension of CVRPTW, where the number of available vehicles is explicitly limited to a prescribed number \( L \). This constraint is common in tactical fleet-planning scenarios where chartering or hiring additional vehicles is either impossible or economically unfeasible. Compared to the classical CVRPTW, the key difference is the explicit upper bound on the number of active routes. This transforms the model from a pure cost-minimization problem to one that also enforces resource scarcity.

\subsubsection{Additional Parameters.}
\( L \in \mathbb{N}_{+} \) represents the maximum number of vehicles that can be dispatched. All other parameters and variables are inherited directly from CVRPTW.

\subsubsection{Mixed-Integer Linear Programming Model (Incremental Constraints).}

\begin{equation}
\min\; Z = \sum_{k=1}^{L} \sum_{i \in V} \sum_{j \neq i} c_{ij} x_{ij}^{k}
\end{equation}
\begin{equation}
\sum_{k=1}^{L} \sum_{j \neq i} x_{ij}^{k} = 1 \quad \forall i \in V_{\!c}
\end{equation}
\begin{equation}
\sum_{j \in V_{\!c}} x_{0j}^{k} \leq 1 \quad \forall k = 1, \dots, L
\end{equation}
\begin{equation}
\sum_{i \in V_{\!c}} x_{i0}^{k} \leq 1 \quad \forall k = 1, \dots, L
\end{equation}
\begin{equation}
\sum_{k=1}^{L} \sum_{j \in V_{\!c}} x_{0j}^{k} \leq L
\end{equation}

The model retains all constraints from the CVRPTW formulation, including flow conservation, capacity limits, time windows, non-negativity, and binary requirements. The additional constraint (34) explicitly limits the number of vehicles departing from the depot to the prescribed fleet size \( L \), thereby capping the number of active routes. This ensures that the limited fleet continues to meet customer demands, vehicle capacity restrictions, and time-window specifications while incorporating resource scarcity into the optimization.

\section{B. Baseline Model Details}

This paper adopts POMO \cite{kwon2020pomo}, PIP, and PIP-D \cite{bi2024learning} as baseline methods. All three methods are built upon the classical Attention Model (AM) \cite{kool2018attention} and have been extended specifically for the Traveling Salesman Problem (TSP), the Vehicle Routing Problem (VRP), and their variants. These models share an identical encoder–decoder architecture grounded in attention mechanisms and are trained end-to-end using the REINFORCE policy-gradient algorithm.

\subsubsection{Model Architecture}

Given an instance  
\[ X = \{x_1, \dots, x_n\} \in \mathbb{R}^{n \times k}, \]  
where \( n \) is the number of nodes and \( k \) is the dimensionality of node features (typically 2-D coordinates or higher-dimensional embeddings), the model follows the encoder-decoder paradigm:  
\[ \begin{aligned}
\{h^{(N)}_1, \dots, h^{(N)}_n\} &= \text{Encoder}(\{x_1, \dots, x_n\}), \\
\{p_1, \dots, p_n\} &= \text{Decoder}(\{h^{(N)}_1, \dots, h^{(N)}_n\}).
\end{aligned} \]

\subsubsection{Encoder}

The encoder consists of a single linear projection followed by \( N \) stacked multi-head attention blocks. Raw features are first embedded into a high-dimensional latent space:  
\[ h^{(0)}_i = \text{FF}(x_i), \quad i = 1, \dots, n. \]  
Each attention block comprises multi-head self-attention (MHA) and a feed-forward network (FFN), enhanced with residual connections and layer normalization (LN). The update rule for layer \( l \) is  
\[ \begin{aligned}
\hat{h}_i &= \text{LN}\left(h^{(l-1)}_i + \text{MHA}\left(h^{(l-1)}_1, \dots, h^{(l-1)}_n\right)\right), \\
h^{(l)}_i &= \text{LN}\left(\hat{h}_i + \text{FFN}(\hat{h}_i)\right), \quad i = 1, \dots, n.
\end{aligned} \]

\subsubsection{Decoder}

The decoder autoregressively outputs a probability distribution over nodes in \( n \) steps. At time step \( t \), its inputs are: 1) the encoder outputs \( \{h^{(N)}_1, \dots, h^{(N)}_n\} \), and 2) the partial tour \( \{\tau_1, \dots, \tau_{t-1}\} \) already selected.

A context embedding \( c_t \) summarizes the current routing state:  
\[ c_t = \left[\bar{h}^{(N)}, h^{(N)}_{\tau_{t-1}}, h^{(N)}_{\tau_1}\right], \]  
where \( \bar{h}^{(N)} = \frac{1}{n} \sum_{i=1}^n h^{(N)}_i \) is the global graph representation. If \( t = 1 \), placeholders (learnable vectors) are used for \( h^{(N)}_{\tau_{0}} \) and \( h^{(N)}_{\tau_{1}} \).

Next, a single-head attention is computed:  
\[ \begin{aligned}
q_t &= W^Q c_t, \\
k_i &= W^K h^{(N)}_i, \quad v_i = W^V h^{(N)}_i, \quad i = 1, \dots, n.
\end{aligned} \]  
The unnormalized logits are  
\[ u_{t,i} = \begin{cases}
\dfrac{q_t^{\top} k_i}{\sqrt{d_k}}, & i \notin \{\tau_1, \dots, \tau_{t-1}\}, \\
-\infty, & \text{otherwise},
\end{cases} \]  
where \( d_k \) denotes the key dimensionality. A softmax function yields the step-\( t \) probability distribution  
\[ p_t = \text{softmax}(u_t). \]  
The decoder outputs the complete node sequence autoregressively, enabling end-to-end path construction.

\subsubsection{Training Procedure}

The Attention Model is trained using REINFORCE and a Greedy Rollout Baseline, requiring no supervised labels; only the tour length serves as the reward signal. For an instance \( s \), the model parameters \( \theta \) induce a path distribution \( p_\theta(\pi|s) \). The objective is to minimize the expected tour length:  
\[ \mathcal{L}(\theta|s) = \mathbb{E}_{\pi \sim p_\theta(\cdot|s)}\left[L(\pi)\right], \]  
where \( L(\pi) \) is the actual cost of tour \( \pi \). An unbiased gradient is obtained via REINFORCE:  
\[ \nabla_\theta \mathcal{L}(\theta|s) = \mathbb{E}_{\pi \sim p_\theta(\cdot|s)}\left[\left(L(\pi) - b(s)\right) \nabla_\theta \log p_\theta(\pi|s)\right], \]  
where \( b(s) \) is the Greedy Rollout Baseline that reduces gradient variance. All methods in this paper follow this training pipeline without further modifications.

\subsection{B.1 POMO}

POMO \cite{kwon2020pomo} builds on the Attention Model (AM) by introducing three key innovations:

\begin{enumerate}
    \item \textbf{Multi-start Exploration.} During training, instead of using a fixed START token, each node is used as the starting point in turn, generating \( N \) parallel complete trajectories per instance. All trajectories share the same network parameters but start from different nodes. This approach incorporates multiple equivalent optimal sequences into the learning signal, significantly enhancing sample efficiency.
    \item \textbf{Shared Baseline.} In the REINFORCE gradient estimator, the average return of all \( N \) trajectories on the same instance is used as a unified baseline. This method requires no additional networks and provides zero-mean, low-variance benefits.
    \item \textbf{Multi-greedy Rollout Inference.} At inference, deterministic greedy decoding is performed once for each admissible starting node, producing \( N \) candidate solutions. The best solution is selected from these candidates. When combined with instance augmentation techniques (e.g., coordinate rotations and reflections), the candidate pool expands to \( N \times K \) without any stochastic sampling, thereby significantly improving solution quality.
\end{enumerate}

\subsection{B.2 PIP \& PIP-D}

PIP enhances POMO by incorporating a one-step lookahead forecast at each decoding step:

\begin{enumerate}
    \item For each unvisited candidate node \( v \), the algorithm tentatively selects \( v \) as the next step and immediately checks whether this decision would render at least one remaining node permanently infeasible according to the problem constraints.
    \item If such an irreversible infeasibility is detected, \( v \) is marked as ``potentially infeasible" and masked for the current step. This check requires only a single forward simulation and incurs minimal computational overhead.
\end{enumerate}

To eliminate the repeated simulation cost, PIP-D introduces an auxiliary decoder (PIP decoder) that learns the masking mapping:

\begin{enumerate}
    \item The PIP decoder shares the encoder with the main policy and appends a lightweight output head. This head outputs the probability of each candidate node being marked as ``potentially infeasible" via a Sigmoid function.
    \item During training, the PIP decoder is optimized using cross-entropy loss against ground-truth labels obtained from the one-step lookahead. At inference, the learned probabilities are thresholded to produce the mask in a single forward pass, significantly reducing the extra training-time overhead.
\end{enumerate}

\section{C. Data Generation Protocols and Solver Configuration}

\subsection{C.1 TSPTW}


Each TSPTW instance consists of \( n \) nodes, including a depot. Each node is defined by a tuple \((x_i, y_i, e_i, l_i)\), where \((x_i, y_i)\) represents the planar coordinates, and \( e_i \) and \( l_i \) specify the earliest and latest allowable arrival times, respectively. The coordinates are randomly sampled from the uniform distribution \(\mathcal{U}[0, 100]^2\). Time windows are generated at three difficulty levels, all ensuring feasibility: 1) \textbf{Easy}: The earliest arrival time \( e_i \) is sampled from \(\mathcal{U}[0, T_N]\), where \( T_N \) is a relaxed estimate of the tour length for an instance of size \( N \). The latest arrival time \( l_i \) is set to \( e_i + T_N \cdot \mathcal{U}[0.5, 0.75] \). 2) \textbf{Medium}: The same scheme as Easy, but with tighter windows: \( l_i = e_i + T_N \cdot \mathcal{U}[0.1, 0.2] \). 3) \textbf{Hard}: A random permutation \(\tau\) of the nodes is generated. For each node \( i \), let \(\psi_i\) be the cumulative distance from the depot along \(\tau\) up to node \( i \). The earliest arrival time \( e_i \) is sampled from \(\mathcal{U}[\psi_i - 50, \psi_i]\), and the latest arrival time \( l_i \) is sampled from \(\mathcal{U}[\psi_i, \psi_i + 50]\), with \(\eta = 50\). The depot has fixed time windows: \( e_0 = 0 \) and \( l_0 = \max_p l_p + \text{dist}(\text{depot}, p) \).

\subsection{C.2 TSPDL}


A TSPDL instance consists of \( n \) nodes, including a depot, with each node described by a tuple \((x_i, y_i, \delta_i, d_i)\). The coordinates \((x_i, y_i)\) are sampled as described previously. Customer demands are set to \(\delta_i = 1\) for all customer nodes, while the depot has \(\delta_0 = 0\). Draft limits are generated based on a standard TSP instance as follows: 1) Compute the total demand \(\Delta = \sum_{j=0}^n \delta_j\). 2) For a fraction \(\sigma\%\) of the nodes, set the draft limits \(d_i\) to be uniformly distributed between \(\delta_i\) and \(\Delta\), i.e., \(d_i \sim \mathcal{U}[\delta_i, \Delta]\). 3) For the remaining nodes, set \(d_i = \Delta\) (no restriction).

The difficulty level is adjusted by the parameter \(\sigma\): \(75\%\) for \textbf{Medium} difficulty and \(90\%\) for \textbf{Hard} difficulty.

\subsection{C.3 CVRPTW}

We directly adopt the open-source generator released with JAMPR \cite{falkner2020learning}.

\subsection{C.4 CVRPTWLV}


Starting from the CVRPTW model, we introduce a tight vehicle limit. Given the node demands \( d_i \) and the vehicle capacity \( D \), the fleet size is set to \( K = \left\lceil \frac{\sum_{i=0}^n d_i}{D} \right\rceil \). This ensures feasibility while significantly tightening the constraints.

\subsection{C.5 Solver Configuration}

In all experiments, LKH-3 is employed uniformly with 10,000 trials for each instance. To mitigate stochastic variance, each instance is solved three times, and the reported objective value is the average of these three runs.

\subsection{C.6 Inference-Time Data Augmentation Strategy}

To thoroughly explore the solution space, we uniformly apply 8× data augmentation during the inference phase \cite{kwon2020pomo}. Specifically, we apply seven reversible geometric transformations to the original coordinates \((x, y)\), generating seven additional equivalent coordinate sets in addition to the original. These transformations are summarized in Table~\ref{tab:aug}.

All listed transformations are orthogonal isometries of the Euclidean plane, preserving pairwise distances and the underlying graph topology. Consequently, the feasibility and optimality of any TSP/VRP solution remain invariant under these mappings. By evaluating the model in parallel on the original input and its seven augmented counterparts, we achieve a significant increase in sampling coverage and final solution quality without introducing additional network parameters.

\begin{table*}[h]
\centering
\begin{tabular}{ccl}
\toprule
\textbf{Index} & \textbf{Transformation} & \textbf{Geometric Interpretation} \\
\midrule
1 & $(1-x,\;y)$ & Reflection across the vertical line $x=0.5$ \\
2 & $(x,\;1-y)$ & Reflection across the horizontal line $y=0.5$ \\
3 & $(1-x,\;1-y)$ & Point reflection (180° rotation) about the center $(0.5,0.5)$ \\
4 & $(y,\;x)$ & Reflection across the principal diagonal $y=x$ \\
5 & $(1-y,\;x)$ & Diagonal reflection followed by vertical reflection \\
6 & $(y,\;1-x)$ & Diagonal reflection followed by horizontal reflection \\
7 & $(1-y,\;1-x)$ & Diagonal reflection followed by central point reflection \\
\bottomrule
\end{tabular}
\caption{Seven reversible geometric transformations constituting the 8× data augmentation at inference.}
\label{tab:aug}
\end{table*}

\section{D. Additional Experiments}

We conduct a comprehensive ablation study on several key design choices in UCPO, including alternative preference relations, scaling factor variants, preference-pairing strategies, sampling density, sample size, and data augmentation.

\subsection{D.1 Alternative Preference Relations}

\subsubsection{Canonical Definition.} The main paper defines the preference relation \(\tau_i \succ \tau_j\) as follows:
\[
\begin{cases}
I(x, \tau_i) = I(x, \tau_j) = 0 \text{ and } f(x, \tau_i) < f(x, \tau_j); \\
I(x, \tau_i) = 0 \text{ and } I(x, \tau_j) = 1; \\
I(x, \tau_i) = I(x, \tau_j) = 1 \text{ and } L(x, \tau_i, \lambda, \mu) < L(x, \tau_j, \lambda, \mu).
\end{cases}
\]
To assess the impact of this design on convergence and solution quality, we systematically investigate the following alternatives.

\subsubsection{Constraint-only Definition (UCPO-C).} Objective values are disregarded for infeasible solutions; the partial order is induced solely by constraint violation. Formally, \(\tau_i \succ \tau_j\) if and only if:
\[
\begin{cases}
I(x,\tau_i)=I(x,\tau_j)=0,\;f(x,\tau_i)<f(x,\tau_j);\\[2pt]
I(x,\tau_i)=0,\;I(x,\tau_j)=1;\\[2pt]
I(x,\tau_i)=I(x,\tau_j)=1,\;L(x,\tau_i,\lambda,\mu)-f(x,\tau_i)\\
\quad\;<\;L(x,\tau_j,\lambda,\mu)-f(x,\tau_j).
\end{cases}
\]

\subsubsection{Primal-only Definition (UCPO-P).} As a dual counterpart to UCPO-C, we let the objective function be the sole criterion irrespective of feasibility:
\[
\begin{cases}
I(x, \tau_i) = I(x, \tau_j) \text{ and } f(x, \tau_i) < f(x, \tau_j); \\
I(x, \tau_i) = 0 \text{ and } I(x, \tau_j) = 1.
\end{cases}
\]
This design explores the trade-off boundary between the primal and the dual problem.

\subsubsection{Dual-only Definition (UCPO-D).} Feasibility labels are discarded; the partial order is determined entirely in the dual space:
\[
\tau_i \succ \tau_j \iff L(x, \tau_i, \lambda, \mu) < L(x, \tau_j, \lambda, \mu).
\]

\subsubsection{Introducing ties (UCPO-T).}
Inspired by TODO \cite{guo2024todo}, we explicitly allow ties in the constrained-preference modeling for two key reasons: First, BOPO notes that adjacent solutions after ranking often differ only marginally, prompting the use of stride sampling to enhance diversity; a similar phenomenon is observed here. Second, for infeasible solutions, Lagrangian values are frequently too close to provide a meaningful distinction, and imposing a preference can introduce noise. Allowing ties helps mitigate this issue.

Specifically, if the absolute difference between the Lagrangian values of two solutions \(\tau_i\) and \(\tau_j\) is within a threshold \(\alpha\), we consider them equivalent (\(\tau_i \equiv \tau_j\)). This leads to the following preference and tie probabilities (where \(L^i = L(x, \tau_i, \lambda, \mu)\) and \(L^j = L(x, \tau_j, \lambda, \mu)\)):
\begin{equation}
    p^*(\tau_i \succ \tau_j \mid x) = \frac{\exp(L^i)}{\exp(L^i) + \phi \cdot \exp(L^j)},
\end{equation}
\begin{multline}
p^*(\tau_i\!\equiv\!\tau_j\mid x)=\\
\frac{\exp(L^i)\exp(L^j)(\phi^{2}-1)}
{\bigl[\exp(L^i)+\phi\exp(L^j)\bigr]\bigl[\exp(L^j)+\phi\exp(L^i)\bigr]}.
\end{multline}
where $\phi = \exp(\alpha)$. The resulting non-tie loss $\mathcal{L}^{(p)}_{\text{T}}$ and tie loss $\mathcal{L}^{(t)}_{\text{T}}$ are
\begin{multline}
\mathcal{L}^{(p)}_{\text{T}}(\theta)=
-\mathbb{E}_{x\sim\mathcal{X}}\frac{1}{|\mathcal{T}|-1}\sum_{\tau\neq\tau_\star}\\[-2pt]
\log\sigma\!\Bigl(\beta^{\text{T}}\bigl[\log\pi_\theta(\tau_\star\!\mid\!x)-
\log\pi_\theta(\tau\!\mid\!x)-\alpha\bigr]\Bigr).
\end{multline}
\begin{multline}
\mathcal{L}^{(t)}_{\text{T}}(\theta)=
-\mathbb{E}_{x\sim\mathcal{X}}\frac{1}{|\mathcal{T}|-1}\sum_{\tau\neq\tau_\star}\\[-2pt]
\log\sigma\!\Bigl[\log\frac{\exp(2\alpha)-1}
{\bigl(1+\exp(\mu+\alpha)\bigr)\bigl(1+\exp(-\mu+\alpha)\bigr)}\Bigr].
\end{multline}

Here, \(\mu = \beta^{\text{T}} \left[ \log \pi_\theta(\tau_{r_\ast} \mid x) - \log \pi_\theta(\tau \mid x) \right]\) and \(\beta^{\text{T}} \equiv \beta^{\text{Dual}}\). Derivations of these losses are omitted as they are beyond the scope of this paper. Empirically, we set \(\alpha = 0.1\) after hyperparameter tuning.

\begin{table}[!ht]
\centering
\renewcommand{\arraystretch}{1.1}
\newcommand{\B}[1]{\textbf{#1}}
\small
\setlength{\tabcolsep}{2pt}
\begin{tabular}{l ccc ccc}
\toprule
& \multicolumn{3}{c}{\textbf{Medium}}
& \multicolumn{3}{c}{\textbf{Hard}} \\
\cmidrule(lr){2-4} \cmidrule(lr){5-7}
\textbf{Method} & Inst. & Obj. & Gap & Inst. & Obj. & Gap \\
\midrule
UCPO-C    & 0.32\% & 13.269 & 1.91\% & \B{1.18\%} & \B{25.743} & \B{0.52\%} \\
UCPO-P    & 100.00\% & -- & -- & 100.00\% & -- & -- \\
UCPO-D    & 0.37\% & \B{13.258} & \B{1.83\%} & 1.49\% & 25.776 & 0.65\% \\
UCPO-T & 0.40\% & 13.260 & 1.84\% & 1.40\% & 25.737 & 0.50\% \\
\cmidrule(lr){1-7}
UCPO      & \B{0.08\%} & 13.262 & 1.86\% & 1.32\% & 25.750 & 0.55\% \\
\bottomrule
\end{tabular}
\caption{Performance on TSPTW-50 (Different preference relations).}
\label{tab:ucpo-pre}
\end{table}

\subsubsection{Results.}
We evaluate the performance on the Medium and Hard datasets of TSPTW-50, with the assessment results presented in Table \ref{tab:ucpo-pre}.

\begin{itemize}
    \item \textbf{UCPO-C} achieves the best overall performance on the Hard subset, with an instance-level infeasibility rate of 1.18\%, an average objective value of 25.743, and an optimality gap of 0.52\%—all of which are the lowest among all compared variants. In contrast, on the Medium subset, its infeasibility rate rises to 0.32\%, which is higher than UCPO’s 0.08\%. This indicates that when the constraints are less diverse, the preference relation induced by primal–dual violations alone is not sufficiently discriminative.
    \item \textbf{UCPO-P}, which relies solely on the primal problem, yields an infeasibility rate of 100.00\% on both subsets. Without dual information, it is unable to assess feasibility, rendering the induced preference relation blind to constraint satisfaction and thus impractical for constrained optimization.
    \item \textbf{UCPO-D}, using only the dual perspective, attains the best objective value on the Medium subset (13.258 vs. UCPO’s 13.262) but suffers from higher infeasibility rates of 0.37\% and 1.49\% on Medium and Hard, respectively. This confirms that a pure dual view can blur the boundary between feasible and near-feasible solutions: a slightly infeasible solution with a very small constraint violation and an excellent objective may incorrectly dominate a truly feasible one.
    \item \textbf{UCPO-T} introduces tie-breaking to mitigate score collisions. On the Hard subset, it improves the objective value slightly (25.737 vs. UCPO’s 25.750) but degrades the infeasibility rate from 1.32\% to 1.40\%. This suggests that the additional tolerance weakens the sharpness of the preference relation and allows more infeasible solutions to survive.
\end{itemize}

Taking both subsets into account, UCPO and UCPO-C demonstrate the most balanced trade-off: UCPO-C excels on the Hard subset, whereas UCPO achieves the lowest infeasibility rate on the Medium subset (0.08\%) and maintains a competitive optimality gap (1.86\% / 0.55\%). Consequently, UCPO is adopted as the default configuration.

\subsection{D.2 Exploration of Scaling Factors}

We investigate several variants of the scaling factors.

\subsubsection{Dual-only scaling (UCPO-$\beta_D$).} In the Dual Exploration Loss, we completely discard the objective component and retain only the constraint-related part of the Lagrangian. This amplifies the influence of constraints on gradient updates:
\begin{equation}
    \beta^{\text{Dual}}_{\text{D}} = \frac{L(x,\tau,\lambda,\mu)-f(x,\tau)}{L(x,\tau_\circ,\lambda,\mu)-f(x,\tau_\circ)}.
\end{equation}

\subsubsection{Primal-only scaling (UCPO-$\beta_P$).} All scaling factors are computed solely from the objective values, thereby emphasizing the primal objective while relegating constraint satisfaction to the preference probabilities:
\begin{equation}
    \beta^{\text{Margin}}_{\text{P}} = \beta^{\text{Primal}}_{\text{P}} = \beta^{\text{Dual}}_{\text{P}} = \frac{f(x,\tau_{\ast})}{f(x,\tau)}.
\end{equation}

\subsubsection{Step-wise indicator for infeasible solutions  (UCPO-$\beta_C$).} Constraint violations are converted into a binary indicator; the gradient weight becomes independent of the magnitude of violation. This substantially increases the model’s focus on achieving feasibility:
\begin{equation}
    \begin{aligned}
\beta^{\text{Margin}}_{\text{C}} &= \frac{f(x,\tau_\ast)}{C}, \qquad \beta^{\text{Dual}}_{\text{C}} &= 1.
\end{aligned}
\end{equation}

\begin{table}[t]
\centering
\renewcommand{\arraystretch}{1.1}
\newcommand{\B}[1]{\textbf{#1}}
\small
\setlength{\tabcolsep}{2.3pt}
\begin{tabular}{l ccc ccc}
\toprule
& \multicolumn{3}{c}{\textbf{Medium}}
& \multicolumn{3}{c}{\textbf{Hard}} \\
\cmidrule(lr){2-4} \cmidrule(lr){5-7}
\textbf{Method} & Inst. & Obj. & Gap & Inst. & Obj. & Gap \\
\midrule
UCPO-$\beta_D$ & 0.35\% & 13.271 & 1.93\% & \B{1.23\%} & \B{25.737} & \B{0.50\%} \\
UCPO-$\beta_P$ & 0.31\% & 13.265 & 1.88\% & 100.00\% & -- & -- \\
UCPO-$\beta_C$ & 100.00\% & -- & -- & 100.00\% & -- & -- \\
\cmidrule(lr){1-7}
UCPO          & \B{0.08\%} & \B{13.262} & \B{1.86\%} & 1.32\% & 25.750 & 0.55\% \\
\bottomrule
\end{tabular}
\caption{Performance on TSPTW-50 (Different scaling factors).}
\label{tab:ucpo-beta}
\end{table}

\subsubsection{Results.} We evaluate the performance of different variants on the Medium and Hard datasets of TSPTW-50, with the results presented in Table \ref{tab:ucpo-beta}.

\begin{itemize}
    \item \textbf{UCPO-$\beta_D$}: This variant re-weights the preference-pair updates by rescaling the dual-exploration loss, enhancing the model's sensitivity to constraint violations. On the Hard subset, it achieves Inst. of 1.23\%, Obj. of 25.737, and Gap of 0.50\%, outperforming UCPO (1.32\%, 25.750, 0.55\%). The larger and more diverse dual slacks on the Hard subset provide a stronger signal for distinguishing feasible from infeasible solutions. On the Medium subset, the gap narrows, with an Inst. of 0.35\%, slightly higher than UCPO's 0.08\%, due to reduced variability in dual information diminishing the benefit of the rescaled loss.
    \item \textbf{UCPO-$\beta_P$}: This variant scales the loss solely based on primal objective differences. On the Medium subset, it performs adequately with an Inst. of 0.31\%, Obj. of 13.265, and Gap of 1.88\%, slightly worse than UCPO's 0.08\%, 13.262, and 1.86\%. However, on the Hard subset, the absence of constraint information leads to catastrophic failure: the infeasibility rate reaches 100.00\%, rendering Obj. and Gap undefined. Without reliable coupling between objective and feasibility, the scaling factor becomes uninformative, allowing large constraint violations to be masked by small objective values, causing severe policy drift toward infeasible regions.
    \item \textbf{UCPO-$\beta_C$}: This variant entirely removes the influence of infeasible pairs by setting their weight to zero. As a result, the model does not receive gradients that would repel it from the infeasible manifold, leading to an infeasibility rate of 100.00\% on both subsets. The loss of continuous guidance across the feasible-infeasible boundary disrupts optimization dynamics and prevents the discovery of feasible basins of attraction.
\end{itemize}

Overall, UCPO maintains the best overall balance, achieving the lowest infeasibility rate on the Medium subset (0.08\%) while sustaining competitive performance on the Hard subset (1.32\%, 25.750, 0.55\%).

\subsection{D.3 Preference-Pair Construction Strategies}

We systematically study three alternative schemes for constructing preference pairs within UCPO. Meanwhile, we also investigated varying levels of sample utilization.

\subsubsection{All-pairs scheme — UCPO$^{\text{Subsets}}$.}
In the main paper, each sampled trajectory is used at most once, except for the designated pivots  
$\tau_\star = \arg\min_{\tau\in\mathcal T^{\!T}} f(x,\tau)$ and  
$\tau_\circ = \arg\min_{\tau\in\mathcal T^{\!F}} L(x,\tau,\lambda,\mu)$.  
While BOPO argues that this level of sample utilization suffices for learning preference relationships, we further probe UCPO’s capacity by exhaustively pairing every sampled solution with every other, yielding  
$|\mathcal T|(|\mathcal T|-1)/2$ preference pairs.  
The corresponding losses are:
\begin{equation}
    \begin{aligned}
\mathcal{L}^{\text{Margin}}&(\theta)=
  -\mathbb{E}_{x\sim\mathcal{X}}
\frac{1}{N_{\mathcal{T}^T,\mathcal{T}^F}}\sum_{\tau_i\in\mathcal{T}^{\!T},\tau_j\in\mathcal{T}^{\!F},\tau_i\succ\tau_j}
\log\sigma\!\Bigl(\\
  &\beta^{\text{Margin}}\bigl[\log\pi_\theta(\tau_i\!\mid\!x)-\log\pi_\theta(\tau_j\!\mid\!x)\bigr]\Bigr),
\end{aligned}
\end{equation}
\begin{equation}
    \begin{aligned}
\mathcal{L}^{\text{Primal}}&(\theta)=
  -\mathbb{E}_{x\sim\mathcal{X}}
\frac{1}{N_{\mathcal{T}^T}}\sum_{\tau_i,\tau_j\in\mathcal{T}^{\!T},\tau_i\succ\tau_j}
\log\sigma\!\Bigl(\\
  &\beta^{\text{Primal}}\bigl[\log\pi_\theta(\tau_i\!\mid\!x)-\log\pi_\theta(\tau_j\!\mid\!x)\bigr]\Bigr),
\end{aligned}
\end{equation}
\begin{equation}
    \begin{aligned}
\mathcal{L}^{\text{Dual}}&(\theta)=
  -\mathbb{E}_{x\sim\mathcal{X}}
\frac{1}{N_{\mathcal{T}^F}}\sum_{\tau_i,\tau_j\in\mathcal{T}^{\!F},\tau_i\succ\tau_j}
\log\sigma\!\Bigl(\\
  &\beta^{\text{Dual}}\bigl[\log\pi_\theta(\tau_i\!\mid\!x)-\log\pi_\theta(\tau_j\!\mid\!x)\bigr]\Bigr),
\end{aligned}
\end{equation}
with  
$N_{\mathcal{T}^T,\mathcal{T}^F}=|\mathcal{T}^T||\mathcal{T}^F|/2$,  
$N_{\mathcal{T}^T}=|\mathcal{T}^T|(|\mathcal{T}^T|-1)/2$,  
$N_{\mathcal{T}^F}=|\mathcal{T}^F|(|\mathcal{T}^F|-1)/2$.  
Activation conditions remain identical to those in UCPO.

\subsubsection{Best-Worst scheme — UCPO$^{\text{B\&W}}$.}  
At the opposite extreme of sample utilization, we retain only the single most informative pair: the best feasible tour versus the worst infeasible tour. Mirroring the SLIM framework \cite{corsini2024self}, we define  
$$
\tau_\star = \arg\min_{\tau\in\mathcal{T}^{\!T}} f(x,\tau), \qquad
\tau_\circ = \arg\max_{\tau\in\mathcal{T}^{\!F}} L(x,\tau,\lambda,\mu),
$$
and train solely on the preference $\tau_\star \succ \tau_\circ$.  
The unified loss becomes  
\begin{equation}
    \begin{aligned}
\mathcal{L}^{\text{B\&W}}&(\theta)=
  -\mathbb{E}_{x\sim\mathcal{X}}
\log\sigma\!\Bigl(\\
  &\beta\bigl[\log\pi_\theta(\tau_\ast\!\mid\!x)-\log\pi_\theta(\tau_circ\!\mid\!x)\bigr]\Bigr),
\end{aligned}
\end{equation}
where $\beta\in\{\beta^{\text{Margin}},\beta^{\text{Primal}},\beta^{\text{Dual}}\}$ is selected according to the provenance of $\tau_\star$ and $\tau_\circ$.

\subsubsection{Argmax-based scheme — UCPO$^{\arg\max}$.}
The main paper drives optimization toward the best solutions via  
$\tau_\star=\arg\min_{\tau\in\mathcal{T}^{\!T}} f(x,\tau)$ and  
$\tau_\circ=\arg\min_{\tau\in\mathcal{T}^{\!F}} L(x,\tau,\lambda,\mu)$.  
To explore the complementary direction, we instead adopt the worst solutions as negative anchors: 
$$
\hat{\tau}_\star = \arg\max_{\tau\in\mathcal{T}^{\!T}} f(x,\tau), \qquad
\hat{\tau}_\circ = \arg\max_{\tau\in\mathcal{T}^{\!F}} L(x,\tau,\lambda,\mu).
$$
The corresponding losses are:
\begin{equation}
    \begin{aligned}
\mathcal{L}^{\text{Margin}}&(\theta)=
  -\mathbb{E}_{x\sim\mathcal{X}}
\frac{1}{|\mathcal{T}^{\!T}|}\sum_{\tau\in\mathcal{T}^{\!T}}
\log\sigma\!\Bigl(\\
  &\beta^{\text{Margin}}\bigl[\log\pi_\theta(\tau\!\mid\!x)-\log\pi_\theta(\hat{\tau}_\circ\!\mid\!x)\bigr]\Bigr),
\end{aligned}
\end{equation}
\begin{equation}
    \begin{aligned}
\mathcal{L}^{\text{Primal}}&(\theta)=
  -\mathbb{E}_{x\sim\mathcal{X}}
\frac{1}{|\mathcal{T}^{\!T}|-1}\sum_{\tau\in\mathcal{T}^{\!T}\setminus\{\hat{\tau}_\star\}}
\log\sigma\!\Bigl(\\
  &\beta^{\text{Primal}}\bigl[\log\pi_\theta(\tau\!\mid\!x)-\log\pi_\theta(\hat{\tau}_\star\!\mid\!x)\bigr]\Bigr),
\end{aligned}
\end{equation}
\begin{equation}
    \begin{aligned}
\mathcal{L}^{\text{Dual}}(\theta)=
  &-\mathbb{E}_{x\sim\mathcal{X}}
\frac{1}{|\mathcal{T}^{\!F}|-1}\sum_{\tau\in\mathcal{T}^{\!F}\setminus\{\hat{\tau}_\circ\}}
\log\sigma\!\Bigl(\\
  &\beta^{\text{Dual}}\bigl[\log\pi_\theta(\tau\!\mid\!x)-\log\pi_\theta(\hat{\tau}_\circ\!\mid\!x)\bigr]\Bigr),
\end{aligned}
\end{equation}

\subsubsection{Others.}  
Preliminary experiments demonstrated that restricting the training signal to only pairs of the form $\tau_i \succ \tau_j$ with $\tau_i\in\mathcal{T}^{\!T}$ and $\tau_j\in\mathcal{T}^{\!F}$—regardless of whether they are generated via UCPO$^{\text{Subsets}}$ or UCPO$^{\arg\max}$—yields inferior performance under both warm-start and cold-start regimes. Consequently, we do not pursue deeper ablations along these restricted lines.

\begin{table}[t]
\centering
\renewcommand{\arraystretch}{1.1}
\newcommand{\B}[1]{\textbf{#1}}
\small
\setlength{\tabcolsep}{2.3pt}   
\begin{tabular}{l ccc ccc}
\toprule
& \multicolumn{3}{c}{\textbf{Medium}}
& \multicolumn{3}{c}{\textbf{Hard}} \\
\cmidrule(lr){2-4} \cmidrule(lr){5-7}
\textbf{Method} & Inst. & Obj. & Gap & Inst. & Obj. & Gap \\
\midrule
UCPO$^{\text{subset}}$  & 2.08\% & 13.429 & 3.14\% &  7.67\% & 25.629 & 0.07\% \\
UCPO$^{\text{B\&W}}$ & \B{0.08\%} & 13.421 & 3.08\% &  \B{1.09\%} & 25.763 & 0.60\% \\
UCPO$^{\arg\max}$  & 4.92\% & 13.569 & 4.22\% & 27.25\% & 26.140 & 2.07\% \\
\cmidrule(lr){1-7} 
UCPO         & \B{0.08\%} & \B{13.262} & \B{1.86\%} &  1.32\% & \B{25.750} & \B{0.55\%}\\
\bottomrule
\end{tabular}
\caption{Performance on TSPTW-50 (Different preference-pair construction strategies).}
\label{tab:pairs}
\end{table}

\subsubsection{Results.}
We evaluate the performance on the Medium and Hard datasets of TSPTW-50, with the assessment results presented in Table \ref{tab:pairs}.

\begin{itemize}
    \item \textbf{UCPO$^{\text{subset}}$}: This variant employs a ``full-pairing” strategy, constructing preference pairs between every feasible and infeasible sample within the same batch. While this maximizes sample utilization, it leads to significant performance degradation. On the Medium subset, the infeasibility rate rises to 2.08\%, the objective value deteriorates to 13.429, and the optimality gap widens to 3.14\%. On the Hard subset, the infeasibility rate reaches 7.67\%, and the optimality gap is 0.07\%, with an objective value of 25.629, which is worse than UCPO’s 25.750. The abundance of pairs introduces numerous misleading gradients, amplifying noise and detracting from the policy’s ability to converge toward the optimal trajectory.
    \item \textbf{UCPO$^{\text{B\&W}}$}: This variant forms a single pair by selecting the best feasible sample and the worst infeasible sample in each batch, achieving the lowest sample utilization among all variants. Despite this, it delivers the best feasibility control: on the Medium subset, the infeasibility rate matches UCPO at 0.08\%, and on the Hard subset, it further drops to 1.09\%, outperforming UCPO’s 1.32\%. The objective values of 13.421 and 25.763 are slightly higher than UCPO, but the significantly lower infeasibility rates demonstrate that a “fewer but cleaner” pairing strategy effectively suppresses policy drift.
    \item \textbf{UCPO$^{\arg\max}$}: This variant guides the policy to move away from the worst sample in the batch while maintaining the same sample efficiency as UCPO. However, its performance is markedly worse across all metrics: on the Medium subset, the infeasibility rate is 4.92\%, the objective value is 13.569, and the optimality gap is 4.22\%. On the Hard subset, the infeasibility rate soars to 27.25\%, the objective value rises to 26.140, and the optimality gap increases to 2.07\%. Since the true optimum is unique (or extremely sparse) in the complete solution space, most samples are sub-optimal. Merely distancing the policy from the worst sample yields an ill-defined optimization direction. Additionally, the best sample in the batch is exploited only once, further diluting the supervisory signal.
\end{itemize}

Overall, neither maximizing nor minimizing sample utilization guarantees performance gains. The key lies in designing an appropriate pairing scheme that effectively steers the policy toward better trajectories. UCPO achieves this balance, delivering the most consistent performance.

\subsection{D.4 Sampling Density}

BOPO \cite{liao2025bopo} employs a grouped-sampling strategy to construct the preference-training set. Given a sampled sequence of \( N \) solutions ranked by preference:
\[
\tau_1 \succ \tau_2 \succ \cdots \succ \tau_N,
\]
only a subset is retained for training:
\[
\tau_1 \succ \tau_{k+1} \succ \cdots \succ \tau_{i\,k+1} \succ \cdots \succ \tau_{\lfloor n/k\rfloor\,k+1},
\]
where \( k = 0, 1, \ldots, \lfloor n/k\rfloor \).

This sparse-sampling approach is motivated by the observation that as the model nears convergence, adjacent solutions in a ranked batch often become highly similar or even identical, providing no meaningful preference signal. By using interval-based selection, redundant samples are effectively reduced, enhancing the model’s ability to discern subtle differences.

We herby explore the impact of varying group sizes by setting \( k \in \{1, 2, 4\} \).

\begin{table}[!ht]
\centering
\renewcommand{\arraystretch}{1.1}
\newcommand{\B}[1]{\textbf{#1}}
\small
\setlength{\tabcolsep}{3.4pt}
\begin{tabular}{l ccc ccc}
\toprule
& \multicolumn{3}{c}{\textbf{Medium}}
& \multicolumn{3}{c}{\textbf{Hard}} \\
\cmidrule(lr){2-4} \cmidrule(lr){5-7}
\textbf{Density} & Inst. & Obj. & Gap & Inst. & Obj. & Gap \\
\midrule
$k{=}1$ & \B{0.08\%} & \B{13.262} & \B{1.86\%} & \B{1.32\%} & \B{25.750} & \B{0.55\%} \\
$k{=}2$ & 0.16\% & 13.262 & 1.86\% & 1.40\% & 25.746 & 0.53\% \\
$k{=}4$ & 0.21\% & 13.265 & 1.88\% & 1.75\% & 25.762 & 0.59\% \\
\bottomrule
\end{tabular}
\caption{Performance on TSPTW-50 (Different sampling density).}
\label{tab:density}
\end{table}

\subsubsection{Results.} 
We assess the model's performance on the Medium and Hard datasets of TSPTW-50, with results presented in Table \ref{tab:density}.

Table \ref{tab:density} shows that as the sampling density \( k \) increases from 1 to 4, sample utilization decreases, leading to a monotonic decline in overall performance. For instance, on the Hard subset, the infeasibility rate rises from 1.32\% to 1.75\%. This trend slightly deviates from BOPO's observations. The discrepancy likely stems from the fact that in constrained optimization, feasible and infeasible solutions exhibit more pronounced objective gaps. Thus, a moderate increase in sample utilization (\( k = 1 \)) provides richer supervisory signals for both feasibility boundaries and objective improvement, enhancing robustness without compromising feasibility.

\subsection{D.5 Sample Size and Data Augmentation}

To broaden the exploration of the solution space during inference, we employ a data augmentation scheme to enhance solution diversity. Preference-based optimization inherently relies on sampling candidate solutions to construct the preference pairs used for training. We systematically study the sampling budget, which directly governs the number of preference pairs exposed to the model and, in turn, affects its expressiveness. Using problem instances with \( n = 50 \) nodes, we set the sampling sizes to 25, 50, and 100 for investigation. Additionally, we ablate the data augmentation component to isolate and evaluate the model’s intrinsic expressive capacity.

\begin{table}[t]
\centering
\renewcommand{\arraystretch}{1.1}
\newcommand{\B}[1]{\textbf{#1}}
\small                    
\setlength{\tabcolsep}{2.3pt}  
\begin{tabular}{l r ccc ccc}
\toprule
 & & \multicolumn{3}{c}{\textbf{Medium}} & \multicolumn{3}{c}{\textbf{Hard}} \\ 
\cmidrule(lr){3-5} \cmidrule(lr){6-8}
\textbf{Aug.} & \textbf{Samples} & Inst. & Obj. & Gap & Inst. & Obj. & Gap \\ 
\midrule
\multirow{3}{*}{\texttt{x8}}
& 25 & 0.38\% & 13.296 & 2.12\% & 1.65\% & 25.774 & 0.64\% \\
& 50 & 0.08\% & 13.262 & 1.86\% & 1.32\% & 25.750 & 0.55\% \\
&100 & \B{0.01\%} & \B{13.235} & \B{1.65\%} & \B{1.06\%} & \B{25.729} & \B{0.46\%} \\
\cmidrule{2-8} 
\multirow{3}{*}{\texttt{x1}}
& 25 & 0.38\% & 13.464 & 3.41\% & 1.65\% & 25.915 & 1.19\% \\
& 50 & 0.08\% & 13.421 & 3.08\% & 1.32\% & 25.862 & 0.98\% \\
&100 & \B{0.01\%} & \B{13.378} & \B{2.75\%} & \B{1.06\%} & \B{25.818} & \B{0.81\%} \\
\bottomrule
\end{tabular}
\caption{Performance on TSPTW-50 (Different sample sizes and data augmentation).}
\label{tab:sample}
\end{table}

\begin{table*}[t]
\centering
\small
\caption{Assets, Licenses, and Their Usage.}
\label{tab:asset}
\renewcommand{\arraystretch}{1.1} 
\begin{tabular}{@{}llll@{}}
\toprule
\textbf{Type} & \textbf{Asset} & \textbf{License} & \textbf{Usage} \\
\midrule
\multirow{6}{*}{Code} & LKH3 \cite{helsgaun2017extension} & Available for academic use & Evaluation \\
     & AM \cite{kool2018attention}   & MIT License                & Revision \\
     & POMO \cite{kwon2020pomo} & MIT License                & Revision \\
     & PIP \cite{bi2024learning} & MIT License                & Revision \\
     & PIP-D \cite{bi2024learning} & MIT License                & Revision \\
     & JAMPR \cite{falkner2020learning} & MIT License                & Generate Dataset \\
\bottomrule
\end{tabular}
\end{table*}

\subsubsection{Results.} 
We assess the model's performance on the Medium and Hard datasets of TSPTW-50, with results presented in Table \ref{tab:sample}. The table demonstrates that increasing trajectory diversity through larger sample sizes and data augmentation consistently enhances both feasibility and solution quality.

\begin{itemize}
    \item \textbf{Sample Size}: For both augmentation factors, increasing the number of trajectories per batch from 25 to 100 reduces the infeasibility rate and narrows the optimality gap. On the Hard subset, the infeasibility rate drops from 1.65\% to 1.06\%, and the optimality gap decreases from 0.64\% to 0.46\%. Similar trends are observed in the Medium subset. A larger sampling pool better delineates the boundary between feasible and infeasible regions, leading to more accurate feasibility and objective value estimates.
    \item \textbf{Data Augmentation}: At any fixed sample size, eight-fold augmentation (×8) outperforms no augmentation (×1). For example, with 50 samples on the Hard subset, the optimality gap is 0.55\% under ×8 compared to 0.98\% under ×1. Augmentation generates structurally diverse samples, enhancing the separation between feasible and infeasible instances and strengthening the learning signal.
\end{itemize}

In summary, increasing the sample budget and incorporating data augmentation expand the coverage of the solution space. Their combined effect leads to superior feasibility and objective performance.

\section{E. Licenses for existing assets}

The assets used in this work are listed in Table \ref{tab:asset} and are all open-source for academic research. We will release our source code under the MIT License.

\end{document}